%% file: 0_full_paper.tex
\documentclass{article}

\PassOptionsToPackage{numbers, compress}{natbib}
\usepackage[preprint]{neurips_2024}

\usepackage[utf8]{inputenc} 
\usepackage[T1]{fontenc}    
\usepackage{hyperref}       

\usepackage{url}            
\usepackage{booktabs}       
\usepackage{amsfonts}       
\usepackage{nicefrac}       
\usepackage{microtype}      
\usepackage{xcolor}         

\usepackage{ulem}
\usepackage{soul}
\usepackage{graphicx}
\usepackage{amsmath}
\usepackage{amssymb}
\usepackage{mathtools}
\usepackage{amsthm}
\usepackage{bbm}
\usepackage{wrapfig}
\usepackage{subfigure}
\usepackage{caption}
\usepackage{subcaption}
\usepackage{algorithm}
\usepackage{algorithmic}
\usepackage{array}
\usepackage{booktabs}

\title{World Models with Hints of Large Language Models for Goal Achieving}

\author{
  Zeyuan Liu$^{1}$\thanks{Equal contribution. $^\dagger$ Corresponding Authors.}
  \And
  Ziyu Huan$^{2}$\footnotemark[1]
  \And
  Xiyao Wang$^{3}$
  \And
  Jiafei Lyu$^{1}$
  \And
  Jian Tao$^{1}$
  \And
  Xiu Li$^{1 \dagger}$
  \And
  Furong Huang$^{3 \dagger}$
  \And
  Huazhe Xu$^{4,5,6 \dagger}$
  \TEST
  $^{1}$Tsinghua Shenzhen International Graduate School, Tsinghua University \\
  $^{2}$The Ohio State University, $^{3}$University of Maryland, College Park \\
  $^{4}$IIIS, Tsinghua University, $^{5}$Shanghai Qi Zhi Institute, $^{6}$Shanghai AI Lab \\
  \texttt{\{liuzeyua22, lvjf20, tj22\}@mails.tsinghua.edu.cn}\\
  \texttt{ziyuhuan.ac@gmail.com}, \texttt{li.xiu@sz.tsinghua.edu.cn}\\
  \texttt{\{xywang, furongh\}@umd.edu}, \texttt{huazhe\_xu@mail.tsinghua.edu.cn}
}

\begin{document}

\maketitle

\input{1_abstract}

\input{2_intro}

\input{3_backg}

\input{4_method}

\input{5_experiments}

\input{6_conclusion}


\bibliography{ref}
\bibliographystyle{plain}

\input{0_full_paper.bbl}
\appendix
\input{icml2024/8_appendix}

\end{document}

%% file: 1_abstract.tex
\begin{abstract}

Reinforcement learning struggles in the face of long-horizon tasks and sparse goals due to the difficulty in manual reward specification. While existing methods address this by adding intrinsic rewards, they may fail to provide meaningful guidance in long-horizon decision-making tasks with large state and action spaces, lacking purposeful exploration. Inspired by human cognition, we propose a new multi-modal model-based RL approach named Dreaming with Large Language Models~(DLLM). DLLM integrates the proposed hinting subgoals from the LLMs into the model rollouts to encourage goal discovery and reaching in challenging tasks. By assigning higher intrinsic rewards to samples that align with the hints outlined by the language model during model rollouts, DLLM guides the agent toward meaningful and efficient exploration. Extensive experiments demonstrate that the DLLM outperforms recent methods in various challenging, sparse-reward environments such as HomeGrid, Crafter, and Minecraft by 27.7\%, 21.1\%, and 9.9\%, respectively.

\end{abstract}


%% file: 2_intro.tex
\section{Introduction}

Reinforcement learning~(RL) is effective when the agents receive rewards that propel them towards desired behaviors~\citep{silver2021reward,ladosz2022exploration}. However, the manual engineering of suitable reward functions presents substantial challenges, especially in complex environments~\cite{xie2023text2reward,dubey2018investigating}. Therefore, solving tasks with long horizons and sparse rewards has long been desired in RL~\cite{Bai_2023, wu2023goal}.

Existing RL methods address this issue by supplementing the extrinsic rewards provided by the environment with an intrinsic reward as an auxiliary objective such as novelty~\citep{burda2019exploration, zhang2019scheduled, zhang2021made}, surprise~\citep{achiam2017surprise}, uncertainty~\citep{bellemare2017distributional, moerland2017efficient, janz2019successor}, and prediction errors~\citep{stadie2015incentivizing, pathak2017curiosity, burda2019large}. 
Nonetheless, there exist scenarios wherein only a limited set of elements possess inherent factors that are truly valuable to the agent's target objective, rendering the exploration of additional aspects inconsequential or potentially detrimental to the overall system performance~\citep{dubey2018investigating, colas2018gep, du2023guiding}. Some recent researches employ large language models (LLMs) to explore new solutions for this issue~\citep{du2023guiding, zhou2023large, zhang2023adarefiner}. Leveraging prior knowledge from extensive corpus data, these methods aim to encourage the exploration of meaningful states. While these approaches have demonstrated impressive results, they depend on querying the LLM for any unknown environmental conditions, which limits their ability to generalize the acquired language information to other steps. Additionally, due to their model-free nature, these approaches cannot capture the underlying relationships between dynamics and language-based hints. They also fail to leverage planning mechanisms or synthetic data generation to enhance sample efficiency.

\begin{figure*}[h]
\vskip 0.2in
\begin{center}
\centerline{\includegraphics[width=1\textwidth]{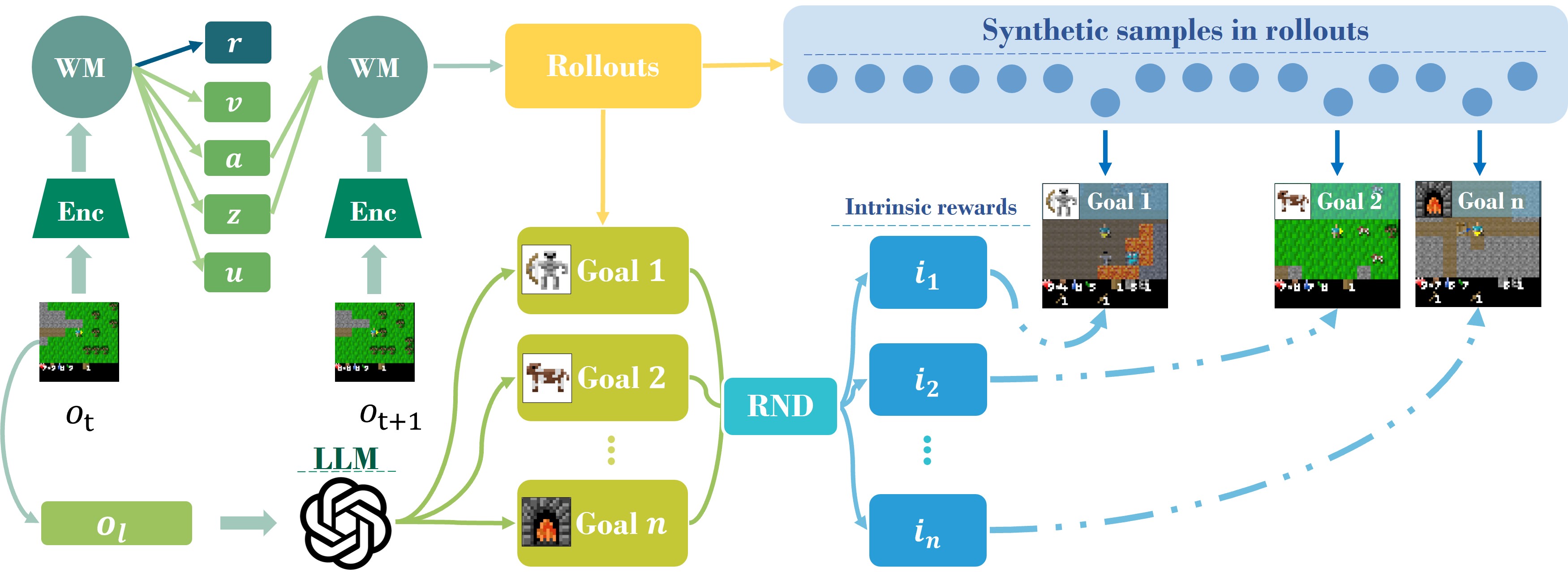}}
\caption{The algorithmic overall structure diagram of DLLM, where WM denotes the world model, $o_l$ represents the natural language caption of the observation, $u$ denotes the transition, and $i_k$ corresponds to the intrinsic reward for the $k$-th goal.}
\label{algorithm_big_figure}
\end{center}
\vskip -0.2in
\end{figure*}

To address this issue, we draw inspiration from how humans solve long-horizon tasks efficiently. Humans excel at breaking down overall goals into several sub-goals and strive to plan a reasonable route to accomplish these goals sequentially~\citep{fernando2018learning}. These goals are often associated with specific actions or environmental dynamics and can ideally be expressed in concise natural language. For example, experienced Minecraft players can naturally connect the action ``obtaining iron'' with its prerequisite actions ``find an iron ore'' and ``breaking iron ore''.

Consequently, we propose Dreaming with Large Language Model (DLLM), a multi-modal model-based RL approach that integrates language hints (i.e., goals) from LLMs into the rollouts to encourage goal discovery and reaching in challenging and sparse-reward tasks, as illustrated in Figure \ref{algorithm_big_figure}. DLLM's world model processes visual inputs and sentence embeddings of natural language descriptions for transitions and learns to predict both. It then rewards the agent when the predicted embeddings are close enough to the goal, facilitating the agents' use of inductive bias to achieve task goals. Thanks to the power of prompt-based LLMs, DLLM can influence agents' behaviors in distinct manners based on the prompts provided for identical tasks, resulting in multiple styles of guidance for the agents. For example, when an agent needs to obtain iron in Minecraft, it can be guided directly to break iron ore, explore more for a better policy, or try interpolating both strategies.

Empirically, we evaluate DLLM on various sparse-reward environments, including Homegrid~\citep{lin2023learning}, Crafter~\citep{hafner2022benchmarking}, and Minecraft~\citep{guss2019minerl}. Experimental results demonstrate that DLLM outperforms recent strong methods in task-oriented and exploration-oriented environments, showcasing robust performance in guiding exploration and training of the agent within highly complex scenarios. In Homegrid, Crafter and Minecraft environments, we successfully improve the performance by \textbf{27.7\%}, \textbf{21.1\%} and \textbf{9.9\%}, respectively, compared to the strongest baseline Dynalang~\citep{lin2023learning}, Achievement Distillation~\citep{moon2024discovering}, and DreamerV3~\citep{hafner2023mastering}. We also observe that leveraging more powerful language models and providing the agent with comprehensive language information results in even better performance.

Our contributions are as follows: (a) we propose DLLM, a multi-modal model-based reinforcement learning approach that utilizes human natural language to describe environmental dynamics, and incorporates LLM's guidance in model rollouts to improve the agent's exploration and goal-completion capabilities; (b) based on goals extracted by LLMs, DLLM can generate meaningful intrinsic rewards through an automatic descending mechanism to guide policy learning; (c) experimental results demonstrate that DLLM outperforms recent strong baselines across diverse environments.

%% file: 3_backg.tex
\section{Background and Related Work}
\textbf{Model-based RL.} Model-based RL (MBRL) trains a world model through online interactions with the environment to predict rewards and next-step states~\cite{silver2016mastering,silver2017mastering,silver2018general}. With the world model, the agent can plan and optimize its policy from imagined sequences~\cite{hansen2022temporal,lowrey2018plan}. Amidst recent advancements, specific contemporary MBRL methods have acquired a world model that is capable of handling high-dimensional observations and intricate dynamics, achieving notable milestones in various domains~\cite{ha2018world,schrittwieser2020mastering,hafner2019dream,hafner2020mastering,hafner2023mastering,hansen2022temporal}. Akin to our approach, the work of Lin et al.~\cite{lin2023learning} constructs a multimodal world model capable of predicting future visual and textual representations, thereby enabling agents to ground their language generation capabilities within an imagined, simulated environment. We employ the same implementation approach, but further integrate the generated natural language from the LLMs during the planning process into constructing the intrinsic rewards.

\textbf{Intrinsically motivated RL.}
In a sparse-reward environment, agents must take many steps in a decision sequence before receiving a positive reward signal. Collecting practical data for training using only random sampling or noisy RL methods is challenging, especially in complex environments with a large state-action space~\cite{reynolds2002reinforcement, yang2021exploration}. Intrinsically motivated RL is the primary method for addressing sparse reward problems. It provides extra intrinsic dense rewards to the agent, encouraging the agent to explore unvisited areas. Pathak et al.~\cite{pathak2017curiosity} propose using curiosity as an intrinsic reward signal, which measures the agent's proficiency in predicting the consequences of its actions within the latent feature space generated by a self-supervised inverse dynamics model. Burda
et al. propose random network distillation (RND)~\cite{burda2018exploration} that leverages the prediction error of a fixed random neural network on novel states and achieves outstanding results in Montezuma's Revenge. Some subsequent studies improve RND via methods such as distributional modeling~\cite{yang2024exploration}. In addition to the intrinsically motivated method that utilizes state novelty, there are methods for maximizing the diversity of states~\cite{teng2012knowledge, linke2020adapting, wang2023coplanner} and for maximizing the diversity of skills mastered by the agent~\cite{baranes2013active, colas2022autotelic}.

Despite the success of intrinsically motivated RL methods, they may face challenges when dealing with large state-action space and complex task scenarios since they are only encouraged to explore novel states, and not all states are useful for achieving the goal. Purposeless exploration can hinder the performance of the agent. Hence, it is essential to incorporate meaningful encouragement to assist the agent, as highlighted in previous studies~\cite{dubey2018investigating, du2023guiding}. This may involve integrating commonsense knowledge, furnishing explicit subgoals as guides, and employing other relevant strategies to facilitate the agent's learning process. DLLM considers these factors with a specific focus on long-term decision-making. During rollouts, DLLM applies intrinsic rewards when the agent achieves goals set by LLM in previous steps, strengthening the understanding of the agent's contextual connections.

\textbf{Leveraging large language models (LLMs) for language goals.}
Pre-trained LLMs showcase remarkable capabilities, particularly in understanding common human knowledge. Naturally, LLMs can generate meaningful and human-recognizable intrinsic rewards for intelligent agents. Choi et al.~\citep{choi2022lmpriors} leverage pre-trained LLMs as task-specific priors for managing text-based metadata within the context of supervised learning. Kant et al.~\citep{kant2022housekeep} utilize LLMs as commonsense priors for zero-shot planning. Similar efforts are made by Yao
et al., Shinn et al., Wu et al., and Wang et al.,~\cite{yao2022react, shinn2023reflexion, wu2023spring, wang2023voyager}, who propose diverse prompt methods and algorithmic structures to mitigate the problems of hallucination and inaccuracy when employing LLMs directly for decision-making. Carta et al.~\citep{carta2023grounding} examine an approach where an agent utilizes an LLM as a policy that undergoes progressive updates as the agent engages with the environment, employing online reinforcement learning to enhance its performance in achieving objectives. Zhang et al.~\citep{zhang2023bootstrap} 
propose to leverage the LLMs to guide skill chaining. Du
et al. propose ELLM~\citep{du2023guiding}, which leverages LLMs to generate intrinsic rewards for guiding agents, integrating LLM with RL. However, the guidance obtained using this approach is only effective in the short term. DLLM draws inspiration from ELLM and endeavors to extend the guidance from LLMs into long-term decision-making.

%% file: 4_method.tex
\section{Preliminaries}
We consider a partially observable Markov decision process (POMDP) defined by a tuple $(S, A, O, \Omega, P, \gamma, R)$, where $s \in S$ represents the states of the environment, $a \in A$ represents the actions, and observation $o \in \Omega$ is obtained from $O(o|s, a)$. $P(s' | s, a)$ represents the dynamics of the environment, $R$ and $\gamma$ are the reward function and discount factor, respectively. During training, the agent’s goal is to learn a policy $\pi$ that maximizes discounted cumulative rewards, i.e., 
$\max\mathbb{E}_{\pi}\left[\sum_{t=0}^{\infty}\gamma^t R(s_t, a_t)\right].$

Additionally, we define two sets of natural language sentence embeddings: the set of sentence embeddings for transitions, denoted as $U$, and the set of sentence embeddings for goals, denoted as $G$. In this context, each $u \in U$ represents a sentence embedding describing the environmental changes from the previous step to the current step, while $g \in G$ represents a sentence embedding of a goal the LLM intends for the agent to achieve. We permit the LLM to output any content within specified formats, thereby enlarging the support of the goal distribution to encompass the space of natural language strings. Thus, $G$ should encompass all possible $u \in U$, i.e., $U \subset G$.

We also define the goal-conditioned intrinsic reward function $R_{\text{int}}(u|g)$, and the DLLM agent optimizes for an intrinsic reward $R_{\rm int}$ alongside the reward $R$ from the environment. Assuming that goals provided by natural language are diverse, common-sense sensitive, and context-sensitive, we expect that maximizing $R_{\text{int}}$ alongside $R$ ensures that the agent maximizes the general reward function $R$ without getting stuck in local optima.

\section{Dreaming with LLMs}
This section systematically introduces how DLLM obtains guiding information (goals) from LLMs and utilizes them to incentivize the agent to manage long-term decision-making.

\subsection{Goal Generation by Prompting LLMs}
To generate the natural language representations of goals and their corresponding vector embeddings, DLLM utilizes a pretrained SentenceBert model~\cite{reimers2019sentence} and GPT~\cite{ouyang2022training}. For GPT, we use two versions including GPT-3.5-turbo-0315 and GPT-4-32k-0315, which we will refer to as GPT-3.5 and GPT-4 respectively in the following text.

We initially obtain the natural language representation, denoted as $o_l$ ($l$ denotes language, and $o_l$ means natural language description of $o$), corresponding to the information in the agent's current observation $o$. This $o_{l}$ may include details such as the agent's position, inventory, health status, and field of view. We use an observation captioner to obtain the $o_l$ following ELLM~\cite{du2023guiding} (see Appendix~\ref{appendix:captioner} for more details of captioners). Subsequently, we provide $o_l$ and other possible language output from the environment (e.g., the task description in HomeGrid) and the description of environmental mechanisms to LLMs to get a fixed number of goals $g^l_{1:K}$ in the form of natural language, where $K$ is a hyperparameter representing the expected number of goals returned by the LLM. We utilize SentenceBert to convert these goals from natural language into vector embeddings $g_{1:K}$. For different environments, we utilize two specific approaches to obtain goals: 1) having the LLM generate responses for $K$ arbitrary types of goals and 2) instructing the LLM to provide a goal for $K$ specified types (e.g., determining which room to enter and specifying the corresponding action). The second approach is designed to standardize responses from the LLM and ensure that the goals output by the LLM cover all necessary aspects for task completion in complex scenarios.

\subsection{Incorporating Decreased Intrinsic Rewards into Dreaming Processes}
At each online interaction step, we have a transition captioner that gives a language description $u_l$ of the dynamics between the observation and the next observation; the language description $u_l$ is then embedded into a vector embedding $u$. Given the sensory representation $x_0$, language description embedding of transition $u_0$, embeddings of goals $g_{1:K}$, and intrinsic rewards for each goal $i_{1:K}$ of replay inputs, the world model and actor produce a sequence of imagined latent states $\hat{s}_{1:T}$, actions $\hat{a}_{1:T}$, rewards $\hat{r}_{1:T}$, transitions $\hat{u}_{1:T}$ and continuation flags $\hat{c}_{1:T}$, where $T$ represents the total length of model rollouts. We use cosine similarity to measure the matching score $w$ between transitions and goals:
\begin{equation}
\begin{aligned}
w\left(\hat{u} \mid g\right)= \begin{cases}\frac{\hat{u} \cdot g}{\left\|\hat{u}\right\|\|g\|} & \text {if }\frac{\hat{u} \cdot g}{\left\|\hat{u}\right\|\|g\|} >M \\ 0 & \text { otherwise }\end{cases},
\end{aligned}
\end{equation}

where $M$ is a similarity threshold hyperparameter. In this step, we aim to disregard low cosine similarities to some extent, thereby preventing misleading guidance. Moreover, within a sequence, a goal may be triggered multiple times. We aim to avoid assigning intrinsic rewards to the same goal multiple times during a single rollout process, as it could lead the agent to perform simple actions repeatedly and eventually diminish the exploration of more complex behaviors. Hence, we only retain a specific goal's matching score when it first exceeds $M$ in the sequence. The method to calculate the intrinsic reward for step $t$ in one model rollout is written as:
\begin{equation}
\begin{aligned}
r_t^{\text {int}} = \alpha \cdot \sum_{k=1}^{K} w_t^k \cdot i_k \cdot \begin{cases} 1 & \text{if} \ \ t_k^{\prime} \ \ \text{exists} \ \ \text{and} \ \ t = t_k^{\prime} \\ 0 & \text{otherwise} \end{cases}
\end{aligned}
\end{equation}
where $\alpha$ is the hyperparameter that controls the scale of the intrinsic rewards, $t'_k$ represents the time step $t$ when the $w^k_t$ first exceeds $M$ within the range of $1$ to $T$.

Then, we give the method to calculate and decrease $i_{1:K}$. If each goal's reward is constant, the agent will tend to repeat learned skills instead of exploring new ones. We use the novelty measure RND~\citep{burda2018exploration} to generate and reduce the intrinsic rewards from LLMs, which effectively mitigates the issue of repetitive completion of simple tasks. To be more specific, after sampling a batch from the replay buffer, we extract the sentence embeddings of the goals from them: $g_{1:B,1:L,1:K}$, where $B$ is the batch size, and $L$ is the batch length. Given the target network $f:G\rightarrow{\mathbb{R}}$ and the predictor neural network $\hat{f_\theta}:G\rightarrow{\mathbb{R}}$, we calculate the prediction error:
\begin{equation}
\begin{aligned}
e_{1:B,1:L,1:K}=\|\hat{f}_{\theta}(g)-f(g)\|^2.
\end{aligned}
\end{equation}
Subsequently, we update the predictor neural network and the running estimates of reward standard deviation, then standardize the intrinsic reward:
\begin{equation}
\begin{aligned}
i_{1:B,1:L,1:K}=(e_{1:B,1:L,1:K}-m)/{\sigma},
\end{aligned}
\end{equation}
where $m$ and $\sigma$ stand for the running estimates of the mean and standard deviation of the intrinsic returns.

\subsection{World Model and Actor Critic Learning}

We implement the world model with Recurrent State-Space Model (RSSM)~\cite{hafner2019learning}, with an encoder that maps sensory inputs $x_t$ (e.g., image frame or language) and $u_{t}$ to stochastic representations $z_t$. Afterward, $z_t$ is combined with past action $a_t$ and recurrent state $h_t$ and fed into a sequence model, denoted as ``seq'', to predict $\hat{z}_{t+1}$:
\begin{equation}
\begin{aligned}
\hat{z}_t, h_t=\operatorname{seq}\left(z_{t-1}, h_{t-1}, a_{t-1}\right),
\end{aligned}
\end{equation}
\begin{equation}
\begin{aligned}
z_t \sim \operatorname{encoder}\left(x_t, u_t, h_t\right),
\end{aligned}
\end{equation}
\begin{equation}
\begin{aligned}
\hat{x}_t, \hat{u}_{t}, \hat{r}_t, \hat{c}_t=\operatorname{decoder}\left(z_t, h_t\right),
\end{aligned}
\end{equation}
where $\hat{z_t}$, $\hat{x}_t$, $\hat{u}_{t}$, $\hat{r}_t$, $\hat{c}_t$ denotes the world model prediction for the stochastic representation, sensory representation, transition, reward, and the episode continuation flag. The encoder and decoder employ convolutional neural networks (CNN) for image inputs and multi-layer perceptrons (MLP) for other low-dimensional inputs. After obtaining multi-modal representations from the decoder and sequence model, we employ the following objective to train the entire world model in an end-to-end manner:
\begin{equation}
\begin{aligned}
\mathcal{L}_{\rm total}={\mathcal{L}_x}+\mathcal{L}_{u}+\mathcal{L}_r+\mathcal{L}_{c}+{\beta_1}\mathcal{L}_{\rm pred}+{\beta_2}\mathcal{L}_{\rm reg},
\end{aligned}
\end{equation}
in which $\beta_1=0.5$, $\beta_2=0.1$, and all sub loss term are written as:
\begin{equation}
\begin{aligned}
&\text{Sensation Loss:} &&\mathcal{L}_x=\left\|\hat{x}_t-x_t\right\|_2^2, \\
&\text{Transition Loss:} &&\mathcal{L}_{u}=\operatorname{catxent}\left(\hat{u}_t, {u}_t\right), \\
&\text{Reward Loss:} &&\mathcal{L}_r=\operatorname{catxent}\left(\hat{r}_t, \operatorname{twohot}\left(r_t\right)\right), \\
&\text{Continue Loss:} &&\mathcal{L}_c=\operatorname{binxent}\left(\hat{c}_t, c_t\right), \\
&\text{Prediction Loss:} &&\mathcal{L}_{\text{pred}}=\max \left(1, \operatorname{KL}\left[\operatorname{sg}\left(z_t\right) \| \hat{z}_t\right]\right), \\
&\text{Regularizer:} &&\mathcal{L}_{\text{reg}}=\max \left(1, \operatorname{KL}\left[z_t \| \operatorname{sg}\left(\hat{z}_t\right)\right]\right),
\end{aligned}
\end{equation}

where $\rm catxent$ is the categorical cross-entropy loss, $\rm binxent$ is the binary cross-entropy loss, $\rm sg$ is the stop gradient operator, KL refers to the Kullback-Leibler (KL) divergence. Details of the $\text{twohot}(\cdot)$ can be found in Appendix \ref{appendix:twohot}.

We adopt the widely used actor-critic architecture for learning policies, where the actor executes actions and collects samples in the environment while the critic evaluates whether the executed action is good. We denote the model state as $s_t = concat(z_t,h_t)$. The actor and the critic give:
\begin{equation}
\begin{aligned}
&\text{Actor: } &&\pi_\theta(a_t \mid s_t), &&\quad &\text{Critic: } && V_\psi(s_t).
\end{aligned}
\end{equation}
Note that both the actor network and the critic network are simple MLPs. The actor aims to maximize the cumulative returns with the involvement of intrinsic reward, i.e.,
\begin{equation}
\begin{aligned}
R_t \doteq \sum_{\tau=0}^{\infty} \gamma^\tau (r_{t+\tau}+r^{\rm int}_{t+\tau}).
\end{aligned}
\end{equation}
The intrinsic rewards beyond the prediction horizon $T$ are inaccessible, and we set them to zero. The details of bootstrapped $\lambda$-returns ~\cite{sutton2018reinforcement}. Then the bootstrapped $\lambda$-returns ~\cite{sutton2018reinforcement} could be written as:
\begin{equation}
\begin{aligned}
R_t^\lambda \doteq r_t+\gamma c_t\left((1-\lambda) V_\psi\left(s_{t+1}\right)+\lambda R_{t+1}^\lambda\right), &&\quad R_T^\lambda \doteq V_\psi\left(s_T\right).
\end{aligned}
\end{equation}

The actor and the critic are updated via the following losses:
\begin{equation}
\begin{aligned}
& \mathcal{L}_V=\operatorname{catxent}\left(V_\psi(s_t), \operatorname{sg}\left(\operatorname{twohot}\left(R_t\right)\right)\right), \\
& \mathcal{L}_\pi=-\frac{\operatorname{sg}\left(R_t-V(s_t)\right)}{\max (1, S)} \log \pi_\theta\left(a_t \mid s_t\right)-\eta \mathrm{H}\left[\pi_\theta\left(a_t \mid s_t \right)\right].
\end{aligned}
\end{equation}

where $S$ is the exponential moving average between the 5th and 95th percentile of $R_t$. We summarize the detailed pseudocode for DLLM in Appendix~\ref{appendix:psecode}. 

%% file: 5_experiments.tex
\section{Experiments}

The primary goal of our experiments is to substantiate the following claim: DLLM helps the agent by leveraging the guidance from the LLM during the dreaming process, thereby achieving improved performance in tasks. Specifically, our experiments test the following hypotheses:
\begin{itemize}
\item \textbf{(H1)} Through proper prompting, DLLM can comprehend complex environments and generate accurate instructions to assist intelligent agents in multi-task environments.
\item \textbf{(H2)} DLLM can leverage the generative capabilities of LLMs to obtain reasonable and novel hints, aiding agents in exploration within challenging environments.
\item \textbf{(H3)} DLLM can significantly accelerate the exploration and training of agents in highly complex, large-scale, near-real environments that necessitate rational high-dimensional planning.
\item \textbf{(H4)} DLLM can be more powerful when leveraging stronger LLMs or receiving additional language information.
\end{itemize}

\textbf{Baselines.} Since we include natural language information in our experiments, we consider employing ELLM~\cite{du2023guiding} and Dynalang~\cite{lin2023learning} as baselines.\footnote{For ELLM and Dynalang, we utilize their official implementations for experimentation. For ELLM, we prompt the LLM and obtain goals following the same procedure as in our method. All language information, including the goals obtained from the LLMs, is encoded into sentence embeddings to feed Dynalang.} We also compare against other recent strong baseline algorithms that do not utilize natural language in each environment.

\textbf{Environments.} We conduct experiments on three environments: HomeGrid~\cite{lin2023learning}, Crafter~\cite{hafner2022benchmarking}, and Minecraft based on MineRL~\cite{guss2019minerl}. These environments span first-person to third-person perspectives, 2D or 3D views, and various types and levels of task complexity.

\textbf{Captioner and Language Encodings.} Within each environment, we deploy an observation captioner and a transition captioner to caption observations and transitions, respectively. Transition captions are stored in the replay buffer for the agent's predictive learning, while observation captions provide pertinent information for LLM. For language encoding, we employ SentenceBert \emph{all-MiniLM-L6-v2}~\cite{wang2020minilm} to convert all natural language inputs into embeddings.

\textbf{The Quality of Generated Goals.} To measure the quality of goals generated by LLMs during online interaction in the environment, we selected the following metrics: novelty, correctness, context sensitivity, and common-sense sensitivity. See the detailed explanations and experiments in Appendix~\ref{appendix:criterias}.

\textbf{Cache.} As each query to the LLM consists of an object-receptacle combination, we implement a cache for each experiment to efficiently reuse queries, thereby reducing both time and monetary cost.

\subsection{HomeGrid}

\textbf{Environment description.} HomeGrid is a multi-task reinforcement learning environment structured as a grid world, where agents get partial pixel observations and language hints (e.g., the descriptions of tasks). We reduce the map size to {10$\times$10} to expedite the overall training of the agent and incorporate icon signals to indicate actions for opening bins. For each step, we have the captioners to caption the observation and transition. The other components of the environment remain unchanged. More details can be seen in Appendix~\ref{appendix:homegridmod}.

To support our claims, we design various settings where the environment provides different levels of information along with distinct language hints for each, as outlined in Table \ref{tab:environment_settings}, to help address hypothesis \textbf{H4}.
\begin{table}[ht]
\centering
\caption{Description of different environment settings}
\begin{tabular}{lp{11.8cm}}
\toprule
\textbf{Setting} & \textbf{Description} \\
\midrule
Standard & The environment provides task descriptions in natural language form. \\
Key info & The environment additionally provides task-relevant objects' location and status information. \\
Full info & The environment additionally provides the location and status information of all objects on the map. For bins, the correct opening actions will also be instructed. \\
Oracle & The agent will always receive accurate instructions. \\
\bottomrule
\end{tabular}
\label{tab:environment_settings}
\end{table}

\textbf{Query prompts, LLM choices, and Goals Generated.} Each query prompt consists of the caption of the agent's current observation and a request for the LLM to generate a goal for each of the two types: ``where to go'' and ``what to do'', respectively. For full prompts and examples, please see Appendix~\ref{appendix:homegridprompt}. We select GPT-4 as the base LLM for all experiments in HomeGrid. Queries to the GPT are made every ten steps. We test the quality of the goals generated in the Appendix~\ref{appendix:homegrid_goals_analysis}.

\begin{wrapfigure}{r}{0.55\textwidth}
\centering
\includegraphics[width=0.55\textwidth]{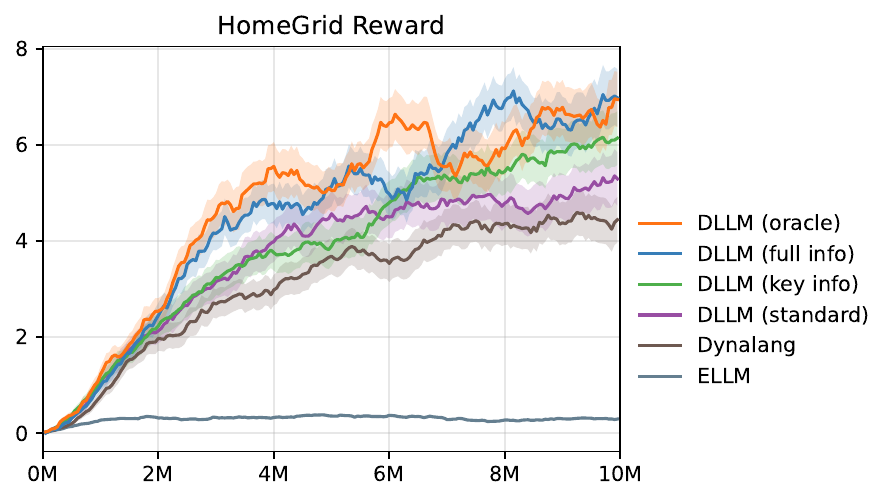}
\caption{HomeGrid experiments results. Curves averaged over 5 seeds with shading representing one-eighth of the standard deviation.}
\label{homegrid}
\vspace{-2mm}
\end{wrapfigure}

\textbf{Performance.} The overall results are depicted in Figure \ref{homegrid}. The baseline algorithm ELLM fails in HomeGrid, likely because it struggles to comprehend the sentence embeddings required to describe the task. Our method outperforms baseline algorithms utilizing the same information in the standard setting, showing strong evidence for \textbf{H1} and \textbf{H3}. Moreover, in the Key info, Full info, and Oracle settings, DLLM demonstrates enhanced performance with increasing information. In the overall context of reduced error prompts in the Full info, DLLM consistently demonstrates a more pronounced advantage throughout the training period. The results of the Full info and Oracle settings show no significant difference. These findings support hypothesis \textbf{H4}.

\subsection{Crafter}

\begin{figure}[ht]
    \centering
    \subfigure[]{
    \includegraphics[width=0.40\linewidth]{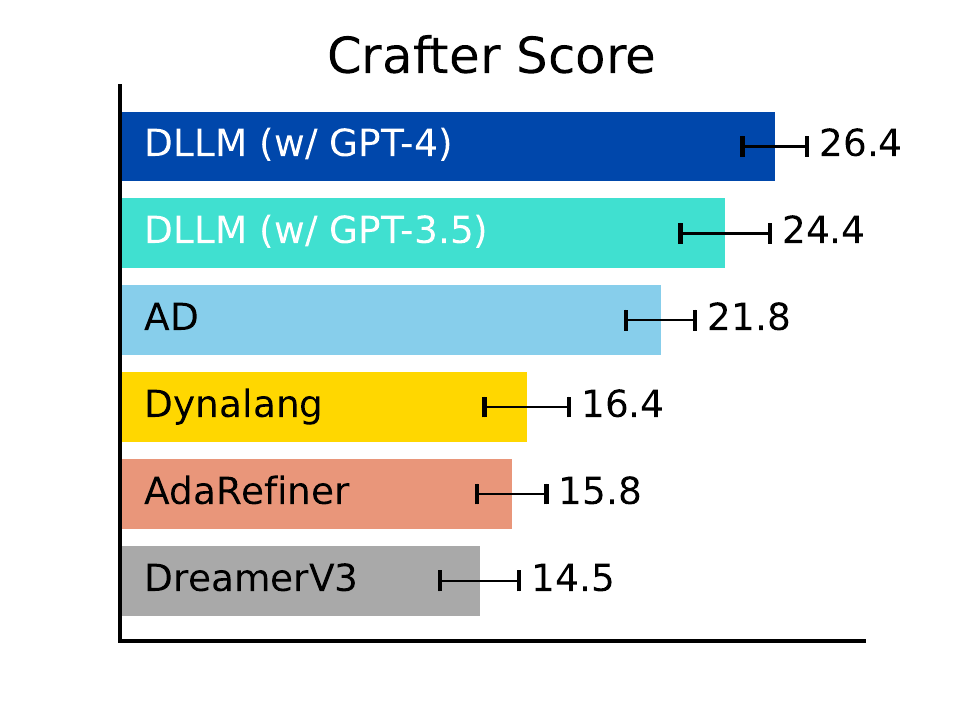}
    \label{fig:crafter_sub_spr}
    }\hspace{0mm}
  \subfigure[]{
    \includegraphics[width=0.56\linewidth]{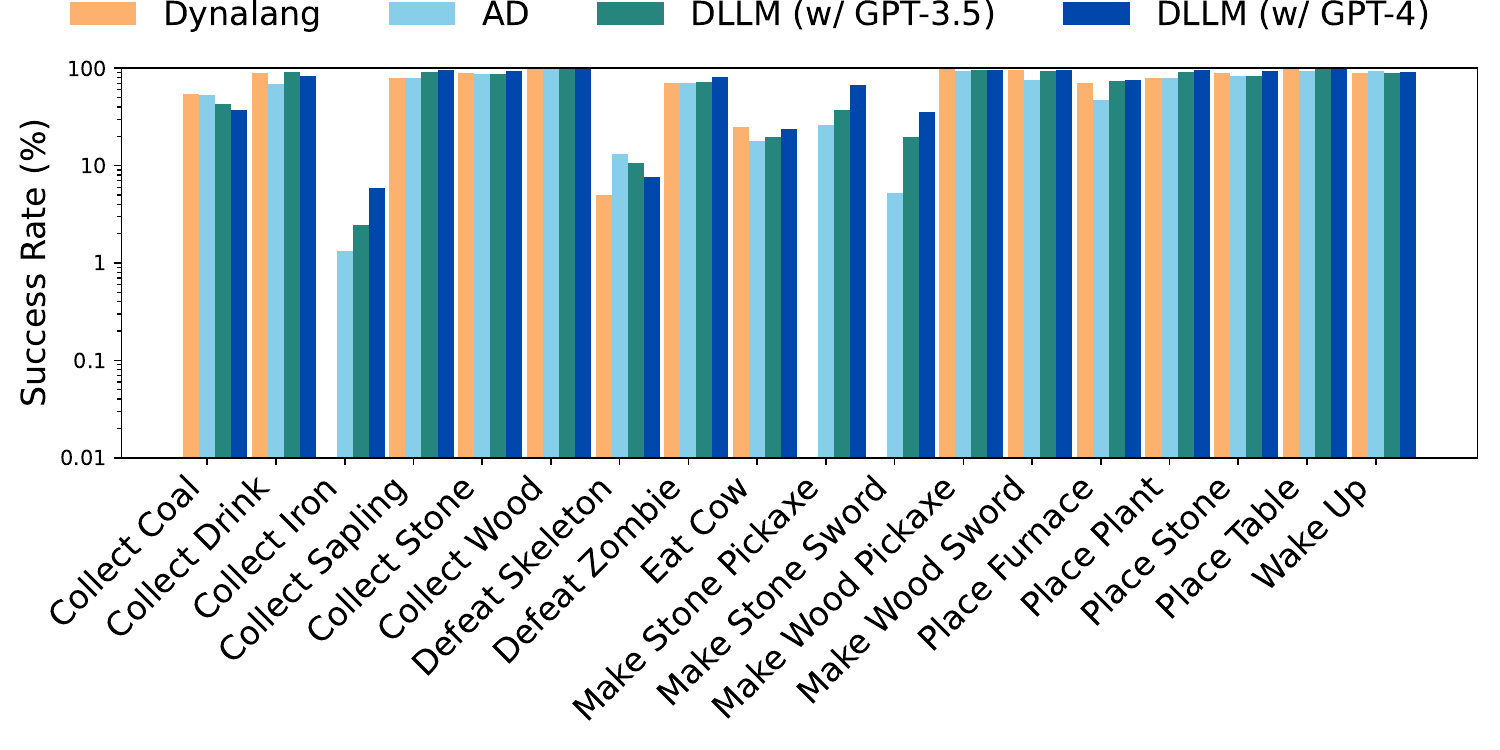}
    \label{fig:crafter_sub_successes}
  }
  \caption{\textbf{Left.} The bar chart comparison of the means and standard deviations between DLLM and baselines. DLLM generally exhibits higher average performance, surpassing baselines by a large margin. \textbf{Right}. The logarithmic scale success rates for unlocking 18 in 22 achievements at 1M (with the remaining four never achieved otherwise). DLLM surpasses baselines in most achievements, particularly excelling in challenging tasks such as ``make stone pickaxe/sword'' and ``collect iron''. ``AD'' refers to Achievement Distillation~\cite{moon2024discovering}, we utilize its official code base to obtain success rate results.}
  \label{crafter_sqr}
  \vspace{-4mm}
\end{figure}

\textbf{Environment description.} The Crafter environment is a grid world that features top-down graphics and discrete action space. Crafter is designed similarly to a 2D Minecraft, featuring a 
procedurally generated, partially observable world where players can collect or craft a variety of artifacts. In Crafter, the player's goal is to unlock the entire achievement tree, which consists of 22 achievements. As the map is designed with entities capable of harming the player (e.g., zombies, skeletons), the player must also create weapons or place barriers to ensure survival.

\textbf{Extra baselines.} We compared three additional types of baselines that do not utilize language information: (1) LLM-based solutions: SPRING~\cite{wu2023spring}, Reflexion~\cite{shinn2023reflexion}, ReAct~\cite{yao2022react}, standalone GPT-4 (step-by-step instructions), (2) model-based RL baseline: DreamerV3~\cite{hafner2023mastering}, (3) model-free methods: Achievement Distillation~\cite{moon2024discovering}, PPO~\cite{schulman2017proximal}, Rainbow~\cite{hessel2017rainbow}. We also add human experts~\cite{hafner2022benchmarking} and random policy as additional references.

\textbf{Query prompts, LLM choices, and Goals Generated.} Each query prompt contains the caption of the agent's current observation description and a request to have the LLM generate five goals. In this portion, we conduct evaluations using two popular LLMs, GPT-3.5 and GPT-4. Through these assessments, we explore whether a more robust LLM contributes to enhanced agent performance, addressing hypothesis \textbf{H4}. Queries to the GPT are made every ten steps. We test the quality of the goals generated in the Appendix~\ref{appendix:crafter_goals_analysis}.

\begin{wraptable}{r}{0.55\textwidth}
    \centering
    \vspace{-4mm}
    \resizebox{\linewidth}{!}{%
    \begin{tabular}{cccc}
      \hline
      Method & Score & Reward & Steps \\
      \hline
      DLLM (w/ GPT-4) & \textbf{38.1}$\pm$1.2 & 15.4$\pm$1.1 & 5M \\
      DLLM (w/ GPT-3.5) & \textbf{37.6}$\pm$1.6 & 14.5$\pm$1.5 & 5M \\
      AdaRefiner (w/ GPT-4) & 28.2$\pm$1.8 & 12.9$\pm$1.2 & 5M \\
      AdaRefiner (w/ GPT-3.5) & 23.4$\pm$2.2 & 11.8$\pm$1.7 & 5M \\
      ELLM & - & 6.0$\pm$0.4 & 5M \\
      \hline
      DLLM (w/ GPT-4) & \textbf{26.4}$\pm$1.3 & 12.4$\pm$1.3 & 1M \\
      DLLM (w/ GPT-3.5) & \textbf{24.4}$\pm$1.8 & 12.2$\pm$1.6 & 1M \\
      Achievement Distillation & 21.8$\pm$1.4 & 12.6$\pm$0.3 & 1M \\
      Dynalang & 16.4$\pm$1.7 & 11.5$\pm$1.4 & 1M \\
      AdaRefiner (w/ GPT-4) & 15.8$\pm$1.4 & 12.3$\pm$1.3 & 1M \\
      PPO (ResNet) & 15.6$\pm$1.6 & 10.3$\pm$0.5 & 1M \\
      DreamerV3 & 14.5$\pm$1.6 & 11.7$\pm$1.9 & 1M \\
      PPO & 4.6$\pm$0.3 & 4.2$\pm$1.2 & 1M \\
      Rainbow & 4.3$\pm$0.2 & 5.0$\pm$1.3 & 1M \\
      \hline
      SPRING (w/ GPT-4) & 27.3$\pm$1.2 & 12.3$\pm$0.7 & - \\
      Reflexion (w/ GPT-4) & 12.8$\pm$1.0 & 10.3$\pm$1.3 & - \\
      ReAct (w/ GPT-4) & 8.3$\pm$1.2 & 7.4$\pm$0.9 & - \\
      Vanilla GPT-4 & 3.4$\pm$1.5 & 2.5$\pm$1.6 & - \\
      \hline
      Human Experts & 50.5$\pm$6.8 & 14.3$\pm$2.3 & - \\
      Random & 1.6$\pm$0.0 & 2.1$\pm$1.3 & - \\
      \hline
    \end{tabular}
    }
    \caption{The results indicate that DLLM with GPT-4 and GPT-3.5 outperforms baseline algorithms, achieving superiority at 1M and 5M training steps.}
    \label{tab:crafter5Mtable_in_main}
    \vspace{-2mm}
\end{wraptable}

\textbf{Performance.} DLLM outperforms all baseline algorithms at 1M and 5M steps. As shown in Figure~\ref{fig:crafter_sub_spr} and Table~\ref{tab:crafter5Mtable_in_main}, DLLM exhibits a significant advantage compared to baselines. Figure~\ref{fig:crafter_sub_successes} shows that DLLM is good at medium to high difficulty tasks like ``make stone pickaxe/sword'' and ``collect iron'' while maintaining stable performance in less challenging tasks. When the steps reach 5M, the performance of DLLM significantly surpasses the language-based algorithm SPRING. These findings show strong evidence for hypotheses \textbf{H2} and \textbf{H3}. In all experiments of Crafter, DLLM (w/ GPT-4) demonstrates a more robust performance than DLLM (w/ GPT-3.5), indicating that DLLM can indeed achieve better performance with the assistance of a more powerful LLM. This finding aligns with the results presented in Appendix~\ref{appendix:crafter_goals_analysis}, where GPT-4 consistently identifies exploration-beneficial goals, thus confirming hypothesis \textbf{H4}. In Crafter, We also include ablation studies on the scale of intrinsic rewards in Appendix~\ref{appendix:crafter_ab_scale}, not decreasing intrinsic rewards in Appendix~\ref{appendix:nornd}, utilizing random goals in Appendix~\ref{appendix:randomgoal}, and allowing repeated intrinsic rewards in Appendix~\ref{appendix:notonepertraj}.

\subsection{Minecraft}

\textbf{Environment description.} Several RL environments, e.g., MineRL~\cite{guss2019minerl}, have been constructed based on Minecraft, a popular video game that features a randomly initialized open world with diverse biomes. Minecraft Diamond is a challenging task based on MineRL, with the primary objective of acquiring a diamond. Progressing through the game involves the player collecting resources to craft new items, ensuring his survival, unlocking the technological progress tree, and ultimately achieving the goal of obtaining a diamond within 36000 steps. In Minecraft Diamond, we also have captioners to provide the captions of the observation and transition in natural language form at each step. The environment settings completely mirror those outlined in DreamerV3~\cite{hafner2023mastering}, which includes awarding a +1 reward for each milestone achieved, which encompasses collecting or crafting a log, plank, stick, crafting table, wooden pickaxe, cobblestone, stone pickaxe, iron ore, furnace, iron ingot, iron pickaxe, and diamond.

\begin{table}[ht]
  \begin{minipage}[b]{0.3\linewidth}
    \centering
    \resizebox{\linewidth}{!}{
    \begin{tabular}{cc}
      \hline
      Method & Reward \\
      \hline
      DLLM (w/ GPT-4) & \textbf{10.0}$\pm$0.3 \\
      DreamerV3 & 9.1$\pm$0.3 \\
      Dynalang & 8.9$\pm$0.4 \\
      IMPALA & 7.4$\pm$0.2 \\
      Rainbow & 6.3$\pm$0.3 \\
      R2D2 & 5.0$\pm$0.5 \\
      PPO & 4.1$\pm$0.2 \\
      ELLM & 0.3$\pm$0.0 \\
      \hline
    \end{tabular}
    }
    \caption{Comparison between DLLM (w/ GPT-4) and baselines in Minecraft at 100M. DLLM (w/ GPT-4) surpasses all baselines, including those that also involve LLMs or natural languages in policy learning.}
    \label{tab:mc_table}
  \end{minipage}\hfill
  \begin{minipage}[b]{0.65\linewidth}
    \centering
    \includegraphics[width=0.9\linewidth]{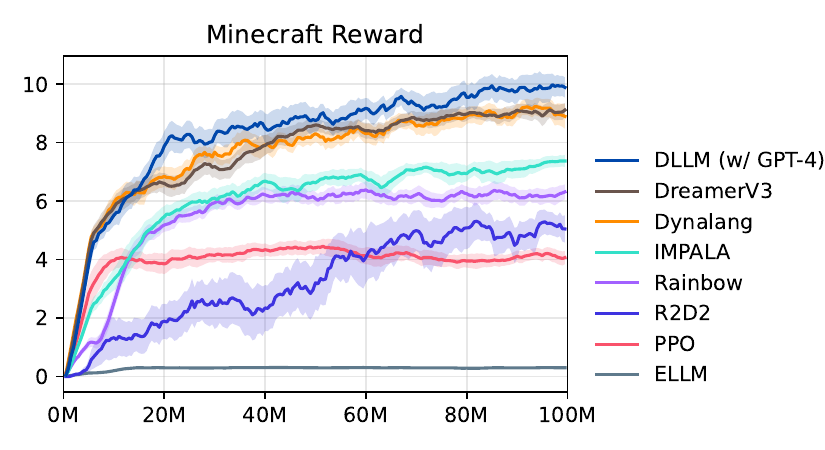}
    \captionof{figure}{The episode returns in Minecraft Diamond. The curves indicate that DLLM enjoys a consistent advantage throughout the entire learning process, thanks to its utilization of an LLM for exploration and training. All algorithms undergo experiments using 5 different seeds.}
    \label{mc_plot}
  \end{minipage}
\end{table}

\textbf{Extra baselines.} To fully compare DLLM with current popular methods from model-based algorithms to model-free algorithms on Minecraft, we include DreamerV3~\cite{hafner2023mastering}, IMPALA~\cite{espeholt2018impala}, R2D2~\cite{liu2003r2d2}, Rainbow~\cite{hessel2017rainbow} and PPO~\cite{schulman2017proximal} as our extra baselines, along with ELLM~\cite{du2023guiding} and Dynalang~\cite{lin2023learning}.

\textbf{Query prompts, LLM choices, and Goals Generated.} During each query to the GPT, we provide it with information about the player's status, inventory, and equipment and request the GPT to generate five goals. We choose GPT-4 as our language model for the DLLM experiment. Please see Appendix \ref{appendix:minecrafterprompt} for specific details. We make a query to the GPT every twenty steps. We also test the quality of the generated goals in Appendix~\ref{appendix:mc_goals_analysis}.

\textbf{Performance.} In Figure~\ref{mc_plot} and Table~\ref{tab:mc_table}, we present empirical results in Minecraft Diamond. Baseline algorithm ELLM struggles in this complex environment, possibly due to high task complexity. DLLM demonstrates higher data efficiency in the early training stages, facilitating quicker acquisition of basic skills within fewer training steps compared to baseline methods. DLLM also maintains a significant advantage in later stages, indicating its ability to still derive reasonable and practical guidance from the LLM during the post-exploration training process. These findings underscore the effectiveness of DLLM in guiding exploration and training in highly complex environments with the support of the LLM, providing compelling evidence for hypothesis \textbf{H3}.

%% file: 6_conclusion.tex
\section{Conclusion and Discussion}

We propose DLLM, a multi-modal model-based RL method that leverages the guidance from LLMs to provide hints (goals) and generate intrinsic rewards in model rollouts. DLLM outperforms recent strong baselines in multiple challenging tasks with sparse rewards. Our experiments demonstrate that DLLM effectively utilizes language information from the environment and LLMs, and enhances its performance by improving language information quality.

\textbf{Limitations.} DLLM relies on the guidance provided by a large language model, making it susceptible to the inherent instability of LLM outputs. This introduces a potential risk to the stability of DLLM's performance, even though the prompts used in our experiments contributed to relatively stable model outputs. Unreasonable goals may encourage the agent to make erroneous attempts, and correcting such misguided behavior may take time. We expect to address these challenges in future work.




%% file: icml2024/8_appendix.tex
\section{Environment Details}

\subsection{HomeGrid}

\subsubsection{Details of Environmental Adjustments}\label{appendix:homegridmod}
\begin{figure}[ht]
    \centering
    \subfigure[]{
    \includegraphics[width=0.46\linewidth]{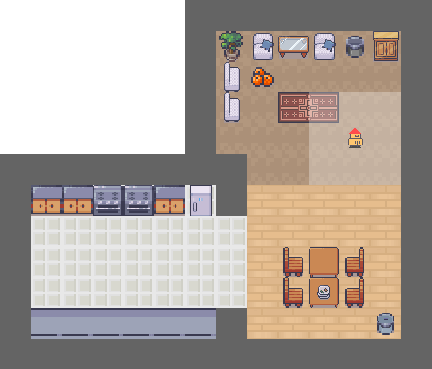}
    \label{fig:homegrid_original}
    }
    \subfigure[]{
    \includegraphics[width=0.31\linewidth]{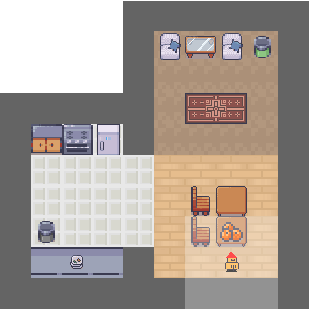}
    \label{fig:homegrid_mini}
    }
    \subfigure[]{
    \includegraphics[width=0.13\linewidth]{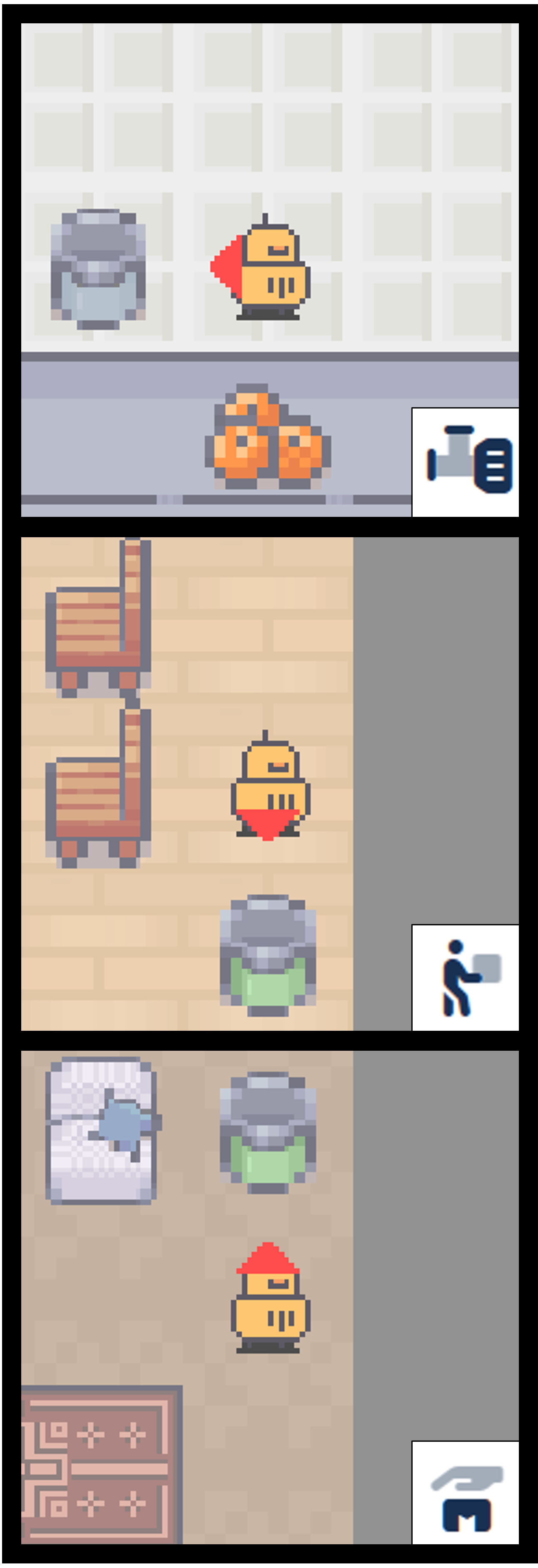}
    \label{fig:homegrid_action}
    }
  \caption{Illustration of the difference between our adapted environment map (b) and the original environment map (a). We generally use a smaller map but ensure that the characteristics of the underlying environment remain unchanged. When and only when the robot succeeds in opening any bin, there will be an icon of the action the robot takes at the lower right of the pixel observation, as shown in (c).}
\end{figure}

``HomeGrid'' is introduced by Dynalang ~\cite{lin2023learning}, and our modified version is based on the ``homegrid-task'' setting. Aside from the pixel observation, this setting additionally provides language information describing the task assigned to the robotic agent. The original map of HomeGrid is a large 14x12 grid, as shown in Figure~\ref{fig:homegrid_original}, and training on such a map would require an excessively long time. We have reduced the map size to a simplified version of 10x10 as in Figure~\ref{fig:homegrid_mini}. In this smaller map, rooms have become more compact, but the width of passages between rooms remains unchanged. In addition to resizing the map, we have adjusted the refresh range for both the player and items, ensuring that players can always move and items can always be accessed. HomeGrid does not provide any visual signal when the robot takes the actions, including ``pedal'', ``lift'', and ``grasp'', representing the different actions to open the bins, so the trained transition captioner needs additional information in the pixel observation. We add icons\footnote{All assets of the icons are collected from \url{https://fontawesome.com/.}} for each of the three actions and make them appear when the related action is taken and the robot succeeds in opening any bin, as shown in Figure~\ref{fig:homegrid_action}. Furthermore, there have been no alterations to HomeGrid's task assignments, reward mechanisms, or total step count.

\subsubsection{Full Prompt Details}\label{appendix:homegridprompt}
During each query to the LLM, we provide the agent with a concise overview of the fundamental aspects of the HomeGrid environment. The observation captioner interprets the current observational state of the environment into natural language, and we provide this to the LLM. We then direct the LLM to choose one goal for ``what to do'' and another for ``where to go.'' In order to ensure consistency in agent responses, we have incorporated mandatory statements and provided illustrative examples. 
GPT's performance can fluctuate, manifesting as inconsistent quality in generated outputs at different times of the day. We recommend capitalizing all the warning text. This can help alleviate the issue.

The actual input provided to the LLM is divided into two parts: system information and game information. The part of system information is:

\framebox[\textwidth]{
  \begin{minipage}{0.95\textwidth} 
      \ttfamily
      You are engaged in a game resembling AI2-THOR. You will receive details about your task, interactive items in view, carried items, and your current room. State the goals you wish to achieve from now on. Please select one thing to do and one room to go, and return them to me, with the format including: \\ \\
      go to the [room], \\
      {[action]} the [object], \\
      {[action]} to [change the status of] (e.g., open) the [bin], \\
      {[action]} the [object] in/to the [bin/room]. \\ \\
      Commas should separate goals and should not contain any additional characters. \\ \\
      An example is: \\
      get the bottle, go to the kitchen.
  \end{minipage}
}

The format for game information is as follows:

\framebox[\textwidth]{
  \begin{minipage}{0.95\textwidth} 
      \ttfamily
      Your task is [text], \\
      You see [objects], \\
      Your carrying is [object], \\
      {[Extra information based on the setting of standard, key info, and full info].}
  \end{minipage}
}

\subsection{Crafter}
Crafter~\cite{hafner2022benchmarking} serves as a platform for reinforcement learning research drawing inspiration from Minecraft, featuring a 2D world where players engage in various survival activities. This game simplifies and optimizes familiar mechanics to enhance research productivity. Players explore a broad world comprising diverse terrains like forests, lakes, mountains, and caves. The game challenges players to maintain health, food, and water, with consequences for neglecting these essentials. The interaction with various creatures, which vary in behavior based on the time of day, adds to the game's complexity.

\subsubsection{Full Prompt Details}\label{appendix:crafterprompt}
During each query to the LLM, we start by presenting the framework of the Crafter environment, employing Minecraft as an analogy. Subsequently, we furnish the current observation information to the LLM, encompassing objects/creatures within the player's field of view, the details of the player's inventory, and the player's status.

In Crafter, we also divide the prompt for the LLM into two sections: system information and game information. The system information is as follows:

\framebox[\textwidth]{
  \begin{minipage}{0.95\textwidth} 
      \ttfamily
      As a professional game analyst, you oversee an RL agent or a player in a game resembling Minecraft. You will receive a starting point that includes information about what the player sees, what the player has in his inventory, and the player's status. For this starting point, please provide the top 5 key goals the player should achieve in the next several steps to maximize its game exploration. \\
      
      Consider the feasibility of each action in the current state and its importance to achieving the achievement. The response should only include valid actions separated by ','. Do not include any other letters, symbols, or words.\\

      An example is: \\
      \texttt{collect wood, place table, collect stone, attack cow, attack zombie.}
  \end{minipage}
}

The format for game information is as follows:

\framebox[\textwidth]{
  \begin{minipage}{0.95\textwidth} 
      \ttfamily
      The player sees [objects/creatures], \\
      The player has [objects], \\
      The status of the player is [text].
  \end{minipage}
}

\subsection{Minecraft}
Minecraft Diamond~\citep{hafner2023mastering} is an innovative environment developed on top of MineRL~\citep{guss2019minerl}, gaining significant attention in the research community within the expansive universe of Minecraft. Minecraft offers a procedurally generated 3D world with diverse biomes, such as forests, deserts, and mountains, all composed of one-meter blocks for player interaction. The primary challenge in this environment is the pursuit of diamonds, a rare and valuable resource found deep underground~\cite{luo2023does}. This quest tests players' abilities to navigate and survive in the diverse Minecraft world, requiring progression through a complex technology tree. Players interact with various creatures, gather resources, and craft items from over 379 recipes, ensuring their survival by managing food and safety.

Developers have meticulously addressed gameplay nuances identified through extensive human playtesting in the Minecraft Diamond environment. Key improvements include modifying the episode termination criteria based on player death or a fixed number of steps and refining the jump mechanism to enhance player interaction and strategy development. The environment, built on MineRL v0.4.415 and Minecraft version 1.11.2, offers a more consistent and engaging experience. The reward system is thoughtfully structured, encouraging players to reach 12 significant milestones culminating in acquiring a diamond. This system, while straightforward, requires strategic planning and resource management, as each item provides a reward only once per episode. The environment's sensory inputs and action space are comprehensive and immersive, offering players a first-person view and a wide range of actions, from movement to crafting.

\subsubsection{Full Prompt Details}\label{appendix:minecrafterprompt}
In Minecraft, we also split the LLM prompt into system and game info sections. The system information is as follows:

\framebox[\textwidth]{
  \begin{minipage}{0.95\textwidth} 
      \ttfamily
      \texttt{As a professional game analyst, you oversee an RL agent or a player in Minecraft, and your final goal is to collect a diamond. You will receive a starting point that includes information about what the player sees, what the player has in his inventory, and the player's status. For this starting point, please provide the top 5 key goals the player should achieve in the next several steps to achieve his final goal.} \\ \\
      \texttt{Take note of the game mechanics in Minecraft; you need to progressively accomplish goals. Each goal should be in the form of an action with an item after it. Please do not add any extra numbers or words.} \\ \\
      \texttt{An example is:} \\
      \texttt{pick up log, attack creepers, drop cobblestone, craft wooden pickaxe, craft arrows.}
  \end{minipage}
}

An example of game information is as follows:

\framebox[\textwidth]{
  \begin{minipage}{0.95\textwidth} 
      \ttfamily
      You have [objects] \\
      You have equipped [objects] \\
      The status of you is [text].
  \end{minipage}
}


\section{Additional Details of DLLM}\label{appendix:additional_details}

\subsection{Two-hot Reward Prediction}\label{appendix:twohot}
We adopt the DreamerV3 approach for reward prediction, utilizing a softmax classifier with exponentially spaced bins. This classifier is employed to regress the two-hot encoding of real-valued rewards, ensuring that the gradient scale remains independent of the arbitrary scale of the rewards. Additionally, we apply a regularizer with a cap at one free nat ~\cite{kingma2016improved} to avoid over-regularization, a phenomenon known as posterior collapse.

\subsection{Pseudo Code}\label{appendix:psecode}
\begin{algorithm}[htb]
   \caption{Dreaming with Large Language Models (DLLM)}
   \label{alg:alg}
\begin{algorithmic}
    \WHILE {acting}
        \STATE Observe in the environment $r_t, c_t, x_t, u_t, o^l_t \leftarrow \operatorname{env}\left(a_{t-1}\right)$.
        \STATE Acquire goals $g^t_{1:K} \leftarrow \text{embed}(\operatorname{LLM}\left(o^l_t\right))$.
        \STATE Encode observations $z_t \sim \operatorname{enc}\left(x_t, u_t, h_t\right)$.
        \STATE Execute action $a_t \sim \pi\left(a_t \mid h_t, z_t\right)$.
        \STATE Add $\left(r_t, c_t, x_t, u_t, a_t, g^t_{1:K}\right)$ to replay buffer.
    \ENDWHILE
    \WHILE {training}
        \STATE Draw batch $\left\{\left(r_t, c_t, x_t, u_t, a_t, g^t_{1:K}\right)\right\}$ from replay buffer.
        \STATE Calculate intrinsic rewards $i_{1:K}$ for each goal using the RND method and update the RND network.
        \STATE Use world model to compute representations ${z}_{t}$, future predictions $\hat{z}_{t+1}$, and decode $\hat{x}_t, \hat{u}_t, \hat{r}_t, \hat{c}_t$.
        \STATE Update world model to minimize $\mathcal{L}_{\rm total}$.
        \STATE Imagine rollouts from all $z_t$ using $\pi$.
        \STATE Calculate match scores $w$ and the intrinsic reward $r^{\rm int}$ for each step.
        \STATE Update actor to minimize $\mathcal{L}_\pi$.
        \STATE Update critic to minimize $\mathcal{L}_V$.
    \ENDWHILE
\end{algorithmic}
\end{algorithm}

\section{Details of Captioners}\label{appendix:captioner}

For the implementation of the captioners, DLLM generally follows ELLM~\cite{du2023guiding}, except that we use trained transition captioners throughout all our experiments to get the language description of the dynamics between two observations. We split the captions into two different parts: semantic parts and dynamic parts.

\subsection{Hard-coded Captioner for Semantic Parts}
The captioner of semantic parts follows the hard-coded captioner implementation outlined in Appendix I of ELLM~\cite{du2023guiding}. The overall semantic captions include the following categories:
\begin{itemize}
    \item \textbf{Field of view.} In the grid world environments (HomeGrid and Crafter), we collect the text description of all the interactable objects in the agent's view, regardless of the object's quantity, to form the caption for this section. Similarly, in the Minecraft environment, we obtain the list of all visible objects from the simulator's semantic sensor.
    \item \textbf{Inventory.} For HomeGrid, this will only include the item the robot carries. For Crafter and Minecraft, we convert each inventory item to the corresponding text descriptor. For Minecraft, we get this information directly from interpreting the observation.
    \item \textbf{Health Status.} In Crafter and Minecraft, if any health statuses are below the maximum, we convert each to a corresponding language description (e.g., we say the agent is ``hungry'' if the hunger status is less than 9). There is no such information in HomeGrid, so we do not provide related captions. Note that the observation directly gives related information in Minecraft, so we simply translate them into natural language.
\end{itemize}

\subsection{Trained Transition Captioner for Dynamics Parts}
The captioner for transitions (dynamics parts) is designed to translate the dynamics between two adjacent observations into natural language form. For convenience, we modify the original simulator to generate language labels for the training of the transition captioner. All language labels use a predetermined and fixed format established by humans. These language labels succinctly describe the dynamics of the environment in the most straightforward manner possible. Notably, these human-designed labels aid the agent in utilizing a similar approach to describe the environment dynamics with concise key words. The designs of all possible formats of language descriptions for transitions in each environment are as follows:

\begin{itemize}
    \item \textbf{HomeGrid.}
        \begin{itemize}
            \item \texttt{go to the [room].}
            \item \texttt{{[action]} the [object]. (e.g., pick up the plates)}
            \item \texttt{{[action]} to [change the status of] (e.g., open) the [bin].}
            \item \texttt{{[action]} the [object] in/to the [bin/room].}
        \end{itemize}
    \item \textbf{Crafter.}
        \begin{itemize}
            \item \texttt{[action] (e.g., sleep, wake up)}
            \item \texttt{[action] the [item/object]. (e.g., attack the zombie)}
        \end{itemize}
    \item \textbf{Minecraft.}
        \begin{itemize}
            \item \texttt{[action] (e.g., forward, jump, sneak)}
            \item \texttt{[action] the [object]. (e.g., craft the torch)}
        \end{itemize}
\end{itemize}

The training process of the captioner mainly follows the methodology outlined in Appendix J of ELLM~\cite{du2023guiding}; we similarly apply a modified ClipCap algorithm~\cite{mokady2021clipcap} to datasets of trajectories generated by trained agents, with details provided in Table \ref{tab:captioner}. Specifically, we embed the visual observations at timestep $t$ and $t+1$ with a pre-trained and frozen CLIP ViT-B-32 model~\cite{radford2021learning}; the embedding is then concatenated together with the difference in semantic embeddings between the corresponding states. Semantic embeddings encompass the inventory and a multi-hot embedding of the set of objects/creatures present in the agent's local view. The concatenated representation of the transition is then mapped through a learned mapping function to a sequence of 32 tokens. We use these tokens as a prefix and decode them with a trained and frozen GPT-2 to generate the caption~\citep{radford2019language}.
\begin{table}[ht]
  \small
  \centering
  \caption{The algorithm used to generate samples, total steps, and scale of the generated dataset for each environment are as follows. We capture one sample every 1K steps during training.}
  \begin{tabular}{cccc}
    \hline
    Environment & Algorithm & Steps & Scale \\
    \hline
    HomeGrid & Dynalang & 10M & 10K \\
    Crafter & Achievement Distillation & 1M & 1K \\
    Minecraft & DreamerV3 & 100M & 100K \\
    \hline
  \end{tabular}
  \label{tab:captioner}
\end{table}

\begin{figure}[ht]
\begin{center}
\centerline{\includegraphics[width=0.90\textwidth]{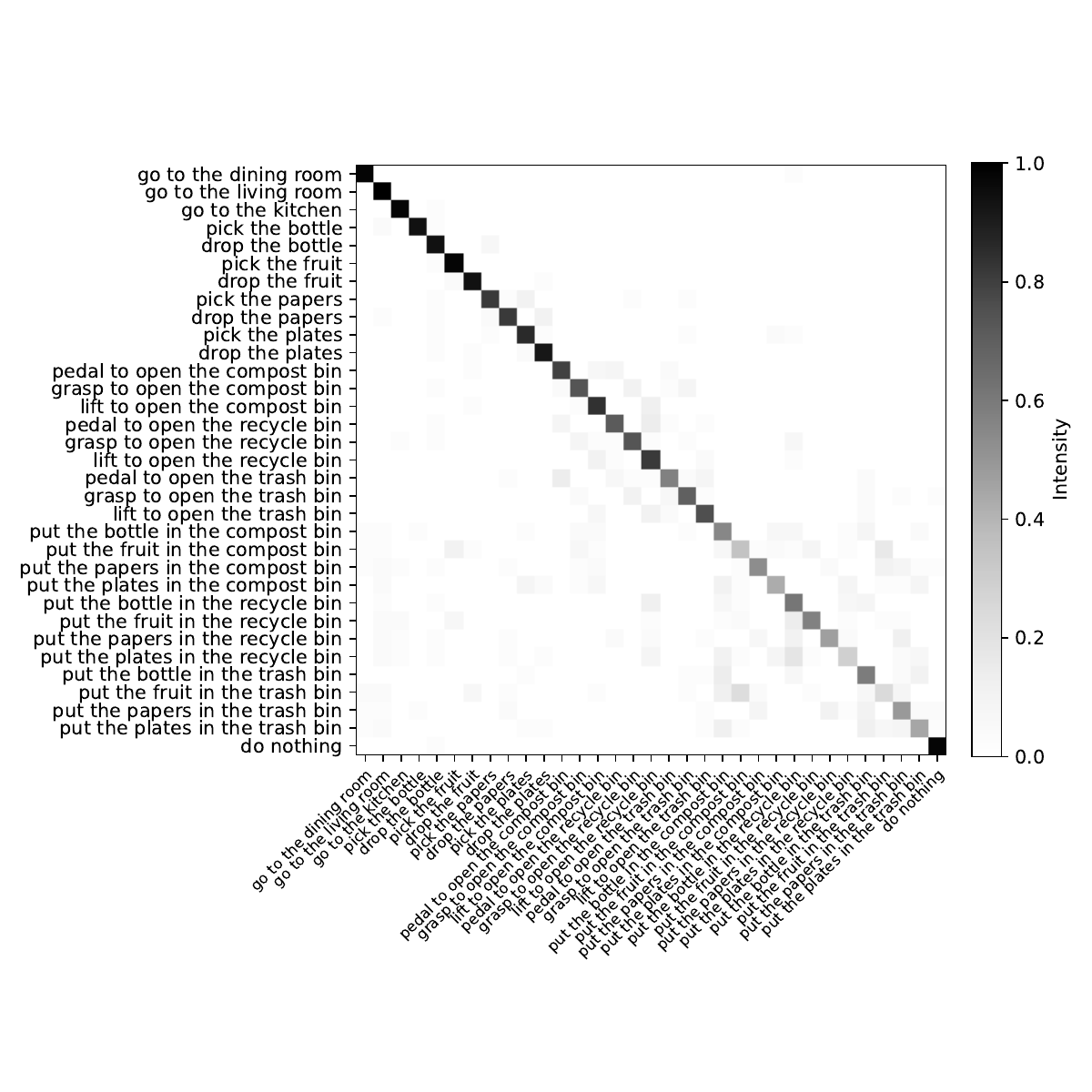}}
\caption{The reward confusion matrix of the trained transition captioner on HomeGrid. Each square's color indicates the probability that the action in the row will be rewarded with the achievement labels on the column. For example, if all action ``go to the dining room'' is recognized as the achievement ``go to the dining room'', we will receive a 100\% on the square corresponding to this row and column. The total in each row does not equal 100\% because multiple rewards may be activated by a single achievement, depending on its description.}
\label{heatmap}
\end{center}
\end{figure}

We employ a reward confusion matrix in Figure~\ref{heatmap} to illustrate the accuracy of our trained transition captioner on HomeGrid, depicting the probability of each achieved goal being correctly rewarded or incorrectly rewarded for another goal during real interactions with the environment. Despite being based on a limited dataset, the captioner demonstrates strong accuracy even when extrapolated beyond the dataset distribution.

\section{Metrics to Test the Quality of Goals Generated by LLMs and Goal Analysis.}\label{appendix:criterias}
Despite the superiority of GPT-3.5 and GPT-4, they may still output impractical or unachievable goals within the game mechanics. This section of ablation experiments primarily investigates the quality of guidance provided by different versions of LLMs in all the environments in which DLLM was conducted. A detailed explanation of the metrics for measuring the generated goals' quality is shown in Table \ref{tab:appendix_criterias}.
\begin{table}[ht]
  \small
  \centering
  \caption{Explanation of the metrics.}
  \begin{tabular}{lp{9.8cm}}
    \toprule
    Metrics & Explanation \\
    \midrule
    Novelty & In exploratory environments like Crafter and Minecraft, a goal is ``novel'' when its prerequisites are fulfilled, but the goal remains unaccomplished in the current episode. For example, in Crafter, the goal ``place table'' is ``novel'' when it is unfulfilled when there are sufficient resources available to achieve it. \\
    Correctness & In task-oriented environments like HomeGrid, each situation has finite correct answers for goals, and other goals may be useless or even lead to task failure. A goal is considered ``correct'' if it is one of the correct answers. For example, ``go to the kitchen'' in HomeGrid is correct if the task is to ``find the papers'' and the papers are located in the kitchen. \\
    Context sensitivity & A goal is ``context sensitivity'' when the player's current field of view and inventory satisfy all the conditions necessary for this goal, regardless of whether these goals are right or novel. e.g., ``make wood pickaxe'' when you have enough resources but already have a wood pickaxe in your inventory, and you see a table. \\
    Common-sense sensitivity & A goal is ``common-sense sensitive''  when it is feasible in the environment in at least one situation. A counterexample is ``make path'' in Crafter, which is impossible. Sometimes, the LLM may not fully understand a previously unknown environment (such as HomeGrid) through concise descriptions, leading to such situations. \\
    \bottomrule
  \end{tabular}
  \label{tab:appendix_criterias}
\end{table}

\subsection{HomeGrid}\label{appendix:homegrid_goals_analysis}
In HomeGrid, given the current observation (including the information about the task and the world state), there is a unique correct answer for ``where to go'' and ``what to do''. To assess the quality of the generated goals, we conducted tests as shown in Table \ref{tab:homegrid_goals}. In this task-oriented environment, we do not test the novelty of goals. The statistical results for each setting were obtained using 1M training samples generated from real interactions. The correctness of goals provided in the standard setting is low since the agent's observation may lack relevant information. There is a noticeable improvement in the Key info setting and extra improvement in the Full info setting. Note that in the Oracle setting, the goals provided to the agent are always correct, so we do not include this setting.
\begin{table}[ht]
  \small
  \centering
  \caption{Testing the quality of goals provided by LLM in each setting of HomeGrid. Ideally, the goals should exhibit high correctness, low context insensitivity, and low commonsense insensitivity.}
  \begin{tabular}{cccc}
    \toprule
    Setting & Standard & Key info & Full info \\
    \midrule
    Correctness & 52.55\% & 61.64\%  &  66.92\%\\
    Context insensitivity & 24.30\% & 18.34\% & 18.95\% \\
    Common-sense insensitivity & 3.45\% & 4.17\% & 5.63\%\\
    \bottomrule
  \end{tabular}
  \label{tab:homegrid_goals}
\end{table}

\subsection{Crafter}\label{appendix:crafter_goals_analysis} 
Given the exploratory nature of the environment, it is hard to say if a goal is ``correct'' or not. Therefore, in evaluating Crafter's goal quality, assessing its correctness holds minimal significance. Instead, our evaluation approach prioritizes novelty over correctness. Through testing various scenarios, the results presented in Table \ref{tab:crafter_goals} indicate that GPT-3.5 tends to offer practical suggestions, demonstrating a context-sensitive ratio of up to 79.41\%. Conversely, GPT-4 leans towards proposing more radical and innovative recommendations, prioritizing novelty. Notably, a goal can exhibit both novelty and context sensitivity concurrently. Therefore, the proportions of ``context insensitivity'' and ``common-sense insensitivity'' in the table are acceptable. Despite GPT-4 showing higher ratios in both context insensitivity and common-sense insensitivity, experimental results underscore its exceptional assistance in enhancing performance. Statistical results for each choice of LLMs were derived from 1M training samples generated from real interactions, with scripts devised to assess these samples without humans in the loop. 

\begin{table}[htb!]
  \small
  \centering
  \caption{Testing the quality of goals provided by GPT-3.5 and GPT-4 in Crafter.}
  \begin{tabular}{cccc}
    \toprule
    LLM & Novelty & Context insensitivity & Common-sense insensitivity \\
    \midrule
    GPT-3.5 & 17.44\% & 20.59\% & 8.26\% \\
    GPT-4 & 38.15\% & 38.80\% & 10.78\% \\
    \bottomrule
  \end{tabular}
  \label{tab:crafter_goals}
\end{table}

\subsection{Minecraft}\label{appendix:mc_goals_analysis}
Despite Minecraft's relative complexity, GPT possesses a wealth of pretrained knowledge about it due to the abundance of relevant information in its training data. Similar to Crafter, correctness is not the primary focus in Minecraft. During the training process of the DLLM, we randomly sampled 1024 steps to collect an equal number of observations, resulting in 5120 goals (1024 multiplied by 5) aligned with the observations. Due to the complexity of elements encompassed within Minecraft, writing scripts to label the quality of goals proves exceedingly challenging. In light of this, we opted for a manual annotation process. This involved a detailed examination of each goal using human labeling. The results are presented in Table~\ref{tab:minecraft_goals}.
\begin{table}[htb!]
  \small
  \centering
  \caption{Testing the quality of goals provided by GPT-4 in Minecraft.}
  \begin{tabular}{cccc}
    \toprule
    Novelty & Context insensitivity & Common-sense insensitivity \\
    \midrule
    73.63\% & 7.66\% & 0.53\% \\
    \bottomrule
  \end{tabular}
  \label{tab:minecraft_goals}
\end{table}

\newpage

\section{Additional Ablation Studies}
\subsection{Token vs Sentence Embedding for Dynalang in HomeGrid}\label{appendix:dynalangtoken}
\begin{figure}[htb!]
\begin{center}
\centerline{\includegraphics[width=0.75\textwidth]{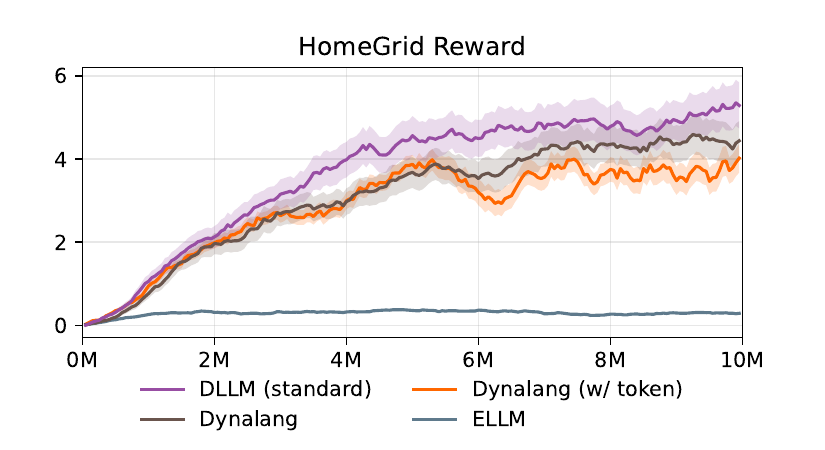}}
\caption{Token vs. sentence embedding performance for Dynalang, averaged across 5 seeds. Dynalang employs a token-by-token approach by tokenizing natural language and passing it into the environment token by token. In contrast, DLLM exclusively utilizes a sentence embedding implementation, as it helps compress a substantial amount of information into a single time step. Within the natural language information we use, task-related language can be separated and still follow Dynalang's token-by-token format. In the HomeGrid environment, we have not observed significant differences.}
\label{plot_ab_dyna}
\end{center}
\end{figure}

This ablation study compares the performance difference of the Dynalang baseline when utilizing token or sentence embedding to acquire natural language information about the task. The results are shown in Figure \ref{plot_ab_dyna}, and we do not observe significant differences between the two methods. Dynalang with token embedding does not outperform Dynalang with sentence embedding. We believe this is because, in our modified environment, Dynalang retrieves task information using tokens and cannot immediately access the complete task information compared to Dynalang using sentence embedding. This is because Dynalang is configured to display only one token per step, requiring time equal to the number of tokens to display all tokens in a sentence.

We do not attempt a similar experiment for natural language information related to transitions and goals because each step in the environment may generate a transition and several goals, and it is impractical to transmit numerous transition tokens token by token to the agent.

\subsection{Ablations of the Intrinsic Reward Scale in Crafter}\label{appendix:crafter_ab_scale}
\begin{figure}[htb!]
    \centering
    \subfigure[Crafter scores.]{
    \includegraphics[width=0.35\linewidth]{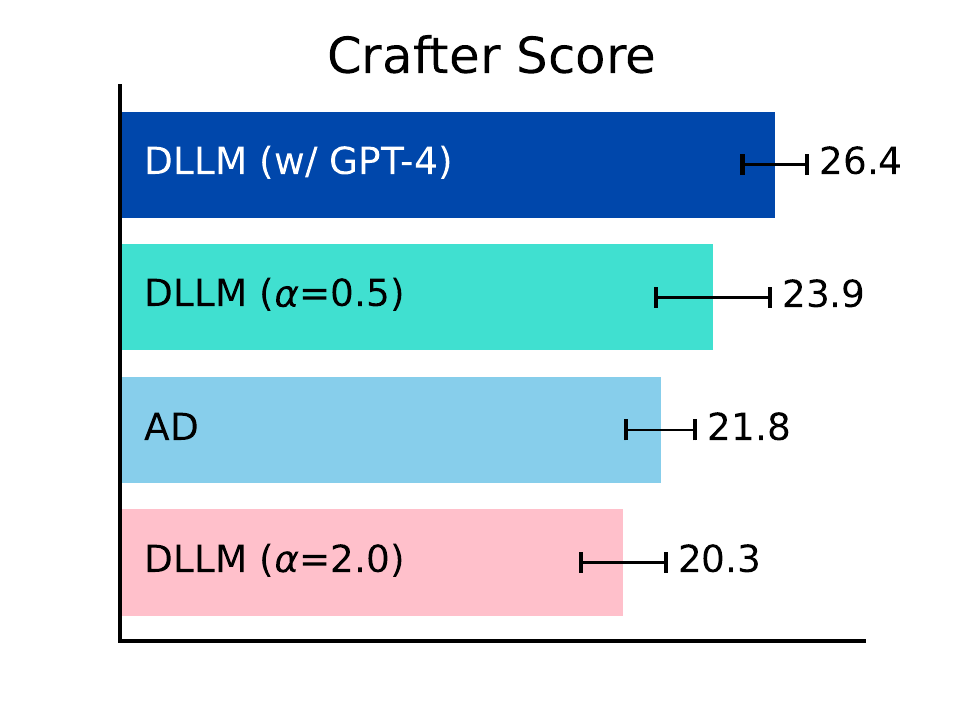}
    \label{fig:Aplot}
    }
    \subfigure[Reward curves.]{
    \includegraphics[width=0.55\linewidth]{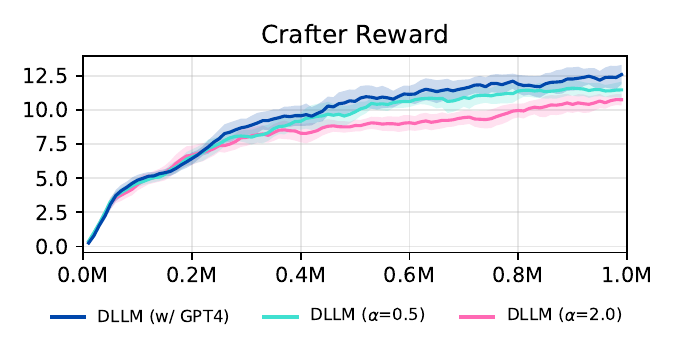}
    \label{fig:Aspr}
    }
    \caption{Experimental results of (a) the mean score values and standard deviations; (b) the reward curves for DLLM with different $\alpha$ comparing against baselines in Crafter, averaged across 5 seeds. ``AD'' refers to Achievement Distillation~\cite{moon2024discovering}.}
\end{figure}

In our work, a Random Network Distillation (RND) network is employed to progressively reduce the intrinsic reward corresponding to each goal. We conduct an ablation experiment to illustrate the necessity of this measure. We set the hyperparameter $\alpha \in$\{0.5, 2\} and perform experiments for each value. $\alpha$ = 2 resulted in catastrophic outcomes, whereas $\alpha$ = 0.5 only led to a slight performance decrease. We conclude that excessively large intrinsic rewards tend to mislead the agent, e.g., try to obtain intrinsic rewards instead of environmental rewards. Conversely, excessively small intrinsic rewards result in inadequate guidance the DLLM provides, undermining its effectiveness in directing the agent's behavior. Please refer to Figures~\ref{fig:Aplot} and~\ref{fig:Aspr} for the results.

\subsection{Decrease or not to decrease intrinsic rewards in Crafter}\label{appendix:nornd}
\begin{figure}[!htb]
    \centering
    \subfigure[Crafter scores.]{
    \includegraphics[width=0.35\linewidth]{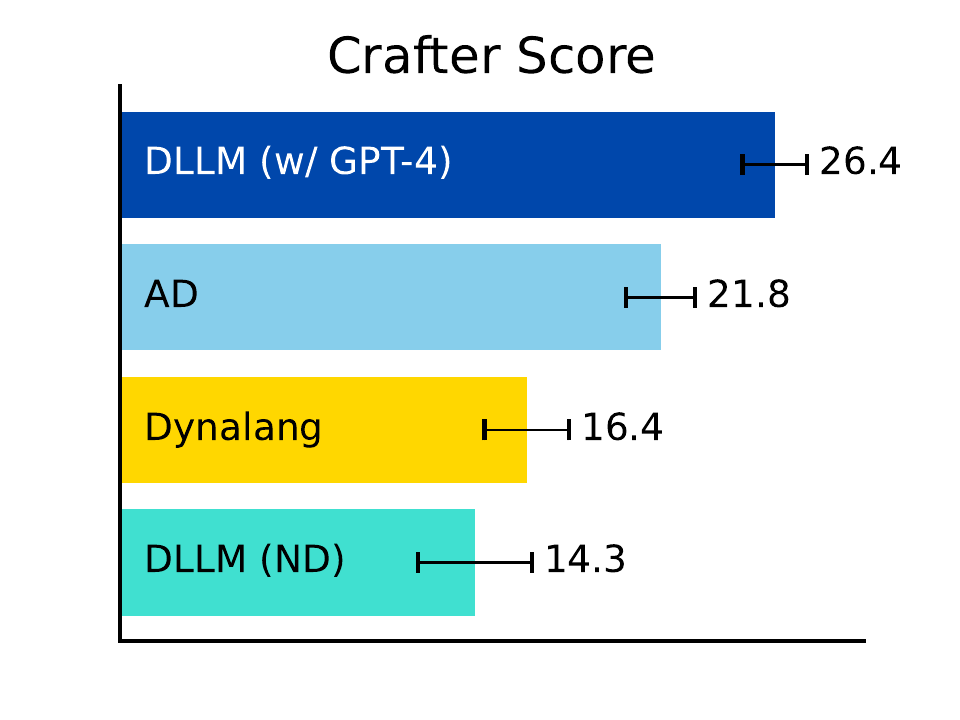}
    \label{fig:norndspr}
    }
    \subfigure[Reward curves.]{
    \includegraphics[width=0.55\linewidth]{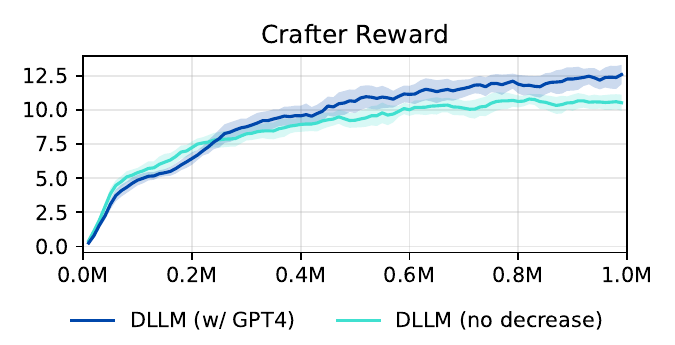}
    \label{fig:norndplot}
    }
   \caption{Experimental results consist of (a) the mean score values and standard deviations; (b) the reward curves for DLLM with decreasing or not decreasing intrinsic rewards, denoted as ``ND'' for ``no decrease'', compared to baselines, averaged across 5 seeds. ``AD'' refers to Achievement Distillation~\cite{moon2024discovering}.}
\end{figure}

This ablation study aims to demonstrate our claim in the paper that repeatedly providing the agent with a constant intrinsic reward for each goal will result in the agent consistently performing simple tasks ~\cite{ riedmiller2018learning, trott2019keeping, devidze2022exploration}, thereby reducing its exploration efficiency and the likelihood of acquiring new skills. We still use an RND network to provide intrinsic rewards in this experiment. However, by preventing the RND network from updating throughout the training process, we ensure that the intrinsic rewards corresponding to all goals remain constant and do not decrease over time. We observe a slight increase in performance during the earlier stages and a significant decline in the later stages, which is consistent with our claim. Please refer to Figures~\ref{fig:norndspr} and~\ref{fig:norndplot} for the results.

\subsection{Random Goals in Crafter}\label{appendix:randomgoal}
\begin{figure}[!htb]
    \centering
    \subfigure[Crafter scores.]{
    \includegraphics[width=0.35\linewidth]{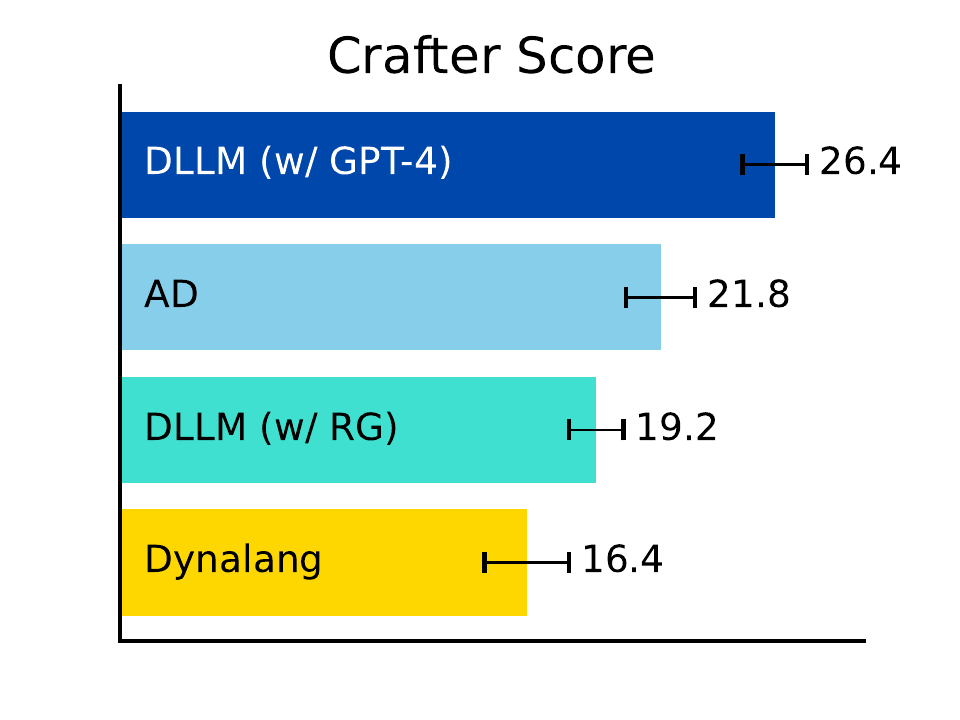}
    \label{fig:randomgoalspr}
    }
    \subfigure[Reward curves.]{
    \includegraphics[width=0.55\linewidth]{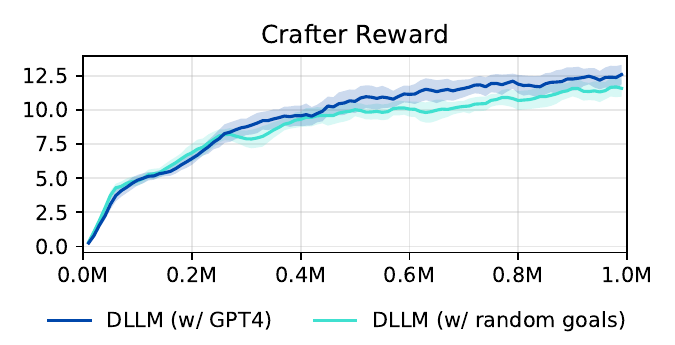}
    \label{fig:randomgoalplot}
    }
   \caption{Experimental results consist of (a) the mean score values and standard deviations; (b) the reward curves for DLLM with random goals, denoted as ``RG'' for ``random goals'', compared to baselines, averaged across 5 seeds. ``AD'' refers to Achievement Distillation~\cite{moon2024discovering}.}
   \label{tab:random_goal_table}
\end{figure}

In this ablation study, we investigate the effectiveness of guidance from the LLM using its pre-trained knowledge compared to randomly sampled goals. In this experiment, we instruct the LLM to sample goals without providing any information about the agent, resulting in entirely random goal sampling. However, we still require the LLM to adhere to the format specified in Appendix\ref{appendix:crafterprompt}. The results are presented in Figure~\ref{fig:randomgoalspr} and \ref{fig:randomgoalplot}. We find that using random goals significantly reduces the performance of DLLM. Nonetheless, DLLM still maintains a certain advantage over recent popular algorithms like Dynalang. This is because providing basic information about the environment to the LLM still generates some reasonable goals in uncertain player conditions. These goals continue to provide effective guidance for the agent through the intrinsic rewards generated in model rollouts.

\subsection{Allow Repetition in Crafter}\label{appendix:notonepertraj}
\begin{figure}[!htb]
    \centering
    \subfigure[Crafter scores.]{
    \includegraphics[width=0.35\linewidth]{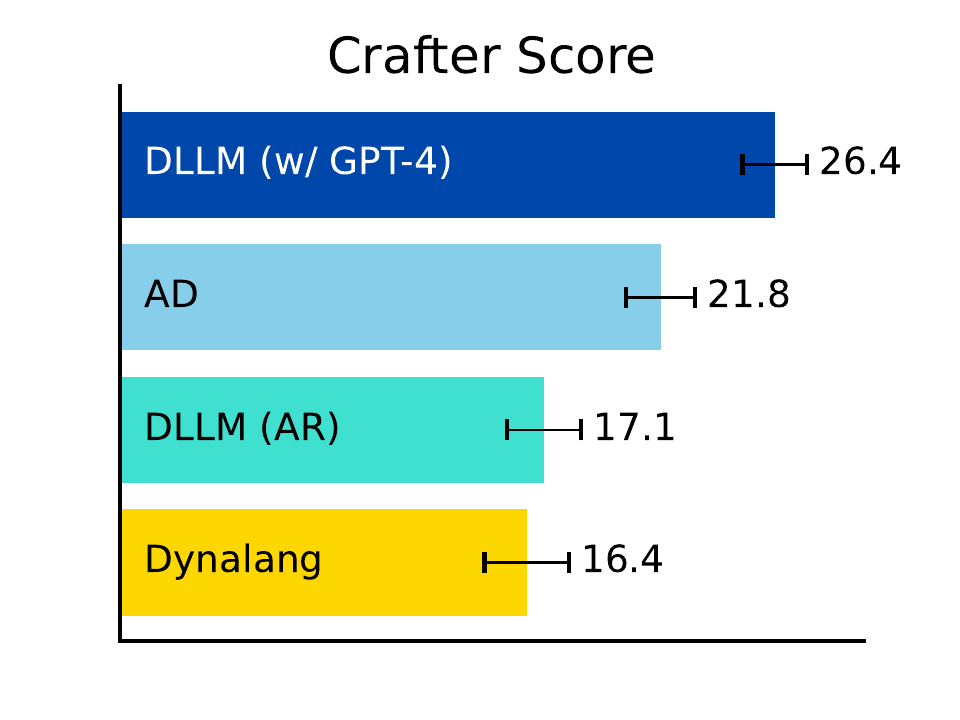}
    \label{fig:notonepertrajspr}
    }
    \subfigure[Reward curves.]{
    \includegraphics[width=0.55\linewidth]{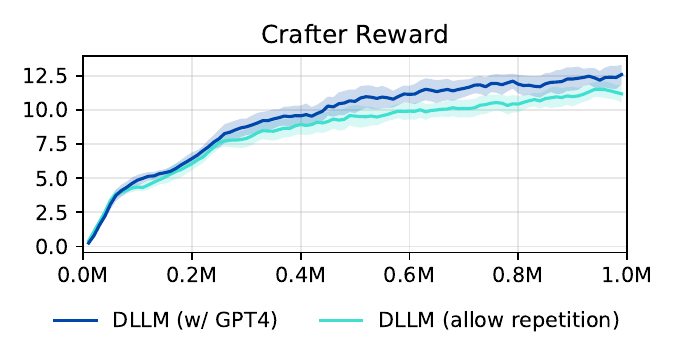}
    \label{fig:notonepertrajplot}
    }
   \caption{Experimental results comprise: (a) the mean score values and standard deviations; (b) the reward curves for DLLM allowing repeated intrinsic rewards for goals, denoted as ``AR'' for ``allow repetition'', compared to baselines, averaged across 5 seeds. ``AD'' refers to Achievement Distillation~\cite{moon2024discovering}.}
   \label{tab:not_one_per_traj_table}
\end{figure}

In Method, we assert that when rewarding the same goal repeatedly within a single model rollout, there is a risk that the agent may tend to repetitively trigger simpler goals instead of attempting to unlock unexplored parts of the technology tree. Consequently, this may lead to decreased performance within Crafter environments primarily focused on exploration. This viewpoint aligns with ELLM~\cite{du2023guiding}. Here, we conducted experiments to substantiate this claim, with results presented in Figure~\ref{fig:notonepertrajspr} and~\ref{fig:notonepertrajplot}. We observed a significant performance decline in DLLM when repetitive rewards for the same goal were allowed.

\newpage
\section{Additional results in Crafter}\label{appendix:5Mcrafter}

\begin{figure*}[!htb]
\vskip 0.2in
\begin{center}
\centerline{\includegraphics[width=0.8\textwidth]{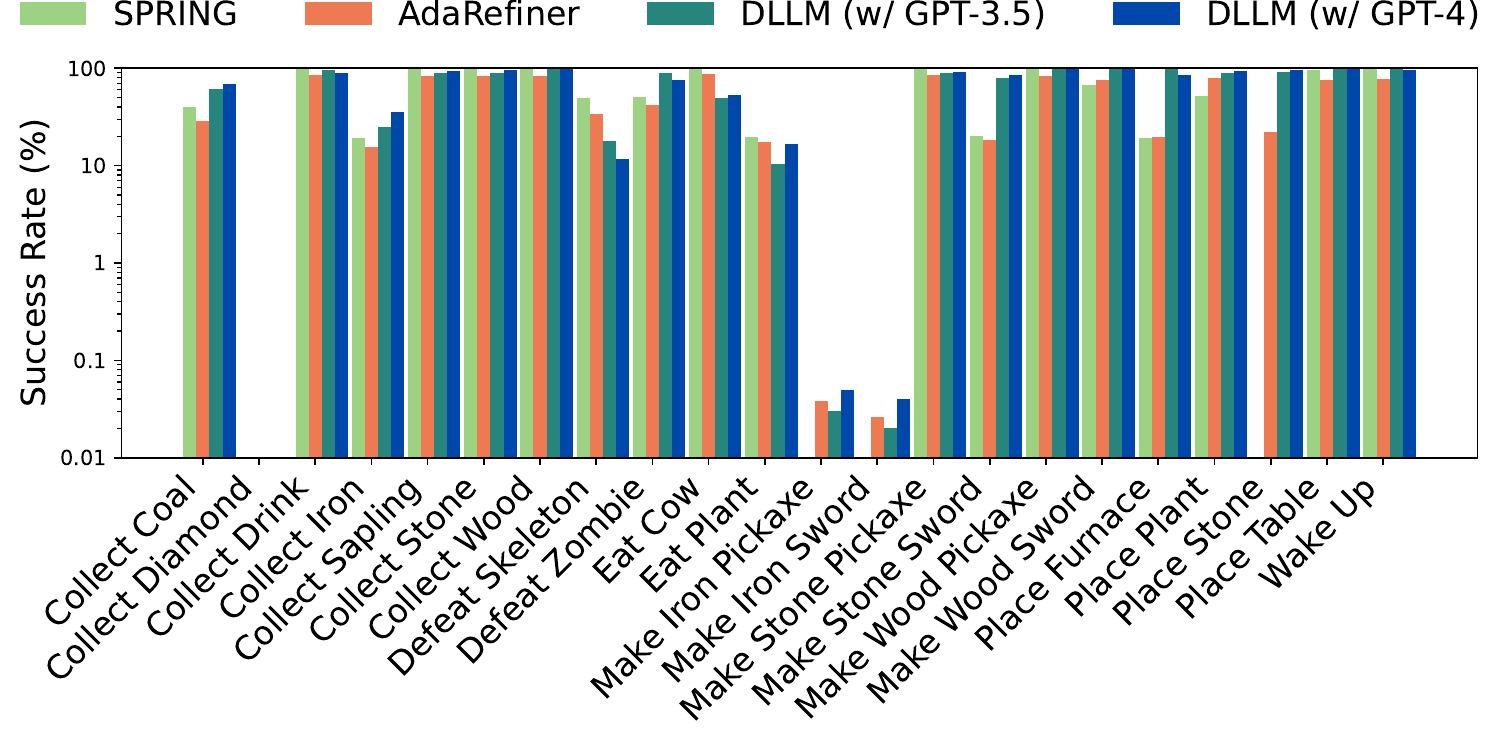}}
\caption{Logarithmic scale success rates for unlocking 22 distinct achievements at 5M steps.}
\label{crafter_sqr5M}
\end{center}
\vskip -0.2in
\end{figure*}

Figure~\ref{crafter_sqr5M} presents the comparison of success rates on the total 22 achievements between DLLM and other baselines in Crafter at 5M steps. DLLM exhibits a higher success rate in unlocking fundamental achievements and outperforms other baselines.

\section{Implementation Details}
For all the experiments, We employ the default hyperparameters for the XL DreamerV3 model~\citep{hafner2024mastering}. Other hyperparameters are specified below. A uniform learning rate of 3e-4 is applied across all environments for the RND networks. Regarding the scale for intrinsic reward $\alpha$, we consistently set $\alpha$ to be 1. We use 1 Nvidia A100 GPU for each single experiment. The training time includes the total GPT querying time, which should be near zero when reusing a cache to obtain the goals.

\begin{table}[ht]
  \small
  \centering
  \caption{Hyperparameters and training information for DLLM.}
    \begin{tabular}{cccc}
        \hline
         & HomeGrid & Crafter & Minecraft \\
        \hline
        Imagination horizon $T$ & 15 & 15 & 15 \\
        Language MLP layers & 5 & 5 & 5 \\
        Language MLP units & 1024 & 1024 & 1024 \\
        Image Size & (64, 64, 3)& (64, 64, 3)& (64, 64, 3)\\
        Train ratio & 32 & 512 & 32 \\
        Batch size & 16 & 16 & 16 \\
        Batch length & 256 & 64 & 64 \\
        GRU recurrent units & 4096 & 4096 & 8192 \\
        Learning rate for RND & 3e-4 & 3e-4 & 3e-4 \\
        The scale for intrinsic rewards $\alpha$ & 1.0 & 1.0 & 1.0 \\
        Similarity threshold $M$ & 0.5 & 0.5 & 0.5 \\
        Max goal numbers $K$ & 2 & 5 & 5 \\
        \hline
        Env steps & 10M & 5M & 100M \\
        Number of envs & 66 & 1 & 64 \\
        Training Time (GPU days) & 2.25 & 10.75 & 16.50 \\
        Total GPT querying Time (days) & 1.50 & 0.75 & 7.50 \\
        \hline
        temperature of GPT & \multicolumn{3}{c}{0.5} \\
        top\_p of GPT  & \multicolumn{3}{c}{1.0} \\
        max\_tokens of GPT  & \multicolumn{3}{c}{500} \\
        \hline
        CPU device & \multicolumn{3}{c}{AMD EPYC 7452 32-Core Processor} \\
        CUDA device & \multicolumn{3}{c}{Nvidia A100 GPU} \\
        RAM & \multicolumn{3}{c}{256G} \\
        \hline
 \end{tabular}
  \label{tab:hyp}
\end{table}

\section{Licenses}
In our code, we have used the following libraries covered by the corresponding licenses:
\begin{itemize}
\item HomeGrid, with MIT license
\item Crafter, with MIT license
\item Minecraft, with Attribution-NonCommercial-ShareAlike 4.0 International
\item OpenAI GPT, with CC BY-NC-SA 4.0 license
\item SentenceTransformer, with Apache-2.0 license
\item DreamerV3, with MIT license
\end{itemize}

\section{Broader Impacts}
LLMs have the potential to produce harmful or biased information. We have not observed LLMs generating such content in our current experimental environments, including HomeGrid, Crafter, and Minecraft. However, applying DLLM in other contexts, especially real-world settings, requires increased attention to social safety concerns. Implementing necessary safety measures involves screening LLM outputs, incorporating restrictive statements in LLM prompts, or fine-tuning with curated data.